\documentclass{article}

\usepackage[preprint]{neurips_2026}

\usepackage[utf8]{inputenc} 
\usepackage[T1]{fontenc}    
\usepackage{hyperref}       
\usepackage{url}            
\usepackage{booktabs}       
\usepackage{amsfonts}       
\usepackage{siunitx} 
\usepackage{nicefrac}       
\usepackage{microtype}      
\usepackage{nicematrix} 
\usepackage{xcolor}         
\usepackage{amsmath}
\usepackage{amssymb}
\usepackage{mathtools}
\usepackage{amsthm}
\usepackage{graphicx}
\usepackage{wrapfig}
\usepackage{lipsum}
\usepackage{color, colortbl}
\usepackage{bbm}
\usepackage{subcaption}
\usepackage{multirow}
\usepackage{tikz}
\usepackage{enumitem}
\usepackage{xspace}
\usepackage[most]{tcolorbox}
\tcbuselibrary{listingsutf8}
\usepackage{makecell}
\usepackage{arydshln}
\newcolumntype{g}{>{\cellcolor{Gray}}c}
\usepackage{subcaption}

\definecolor{tableofcontent}{HTML}{E63E15}
\definecolor{urlcol}{HTML}{2470D8}
\usepackage[normalem]{ulem}
\useunder{\uline}{\ul}{}
\hypersetup{
    colorlinks=true,       
    linkcolor=tableofcontent, 
    urlcolor=black,           
}

\newcommand{\pgftextcircled}[1]{
    \setbox0=\hbox{#1}%
    \dimen0\wd0%
    \divide\dimen0 by 2%
    \begin{tikzpicture}[baseline=(a.base)]%
        \useasboundingbox (-\the\dimen0,0pt) rectangle (\the\dimen0,1pt);
        \node[circle,draw,outer sep=0pt,inner sep=0.1ex] (a) {#1};
    \end{tikzpicture}
}
\setlist{leftmargin=10mm}
\definecolor{Gray}{gray}{0.9}
\newcommand{\xhdr}[1]{{\vspace{1pt}\noindent\bfseries #1}.}
\newcommand{\ie}{\textit{i.e., }}
\newcommand{\eg}{\textit{e.g., }}

\usepackage[most]{tcolorbox}
\usepackage{lipsum}
\tcbuselibrary{skins}
\usepackage{mdframed}
\definecolor{niceblue}{HTML}{3c9dfd}
\usepackage{fontawesome5}
\usepackage{pifont}
\definecolor{light-primary}{RGB}{234, 242, 255}
\definecolor{dark-secondary}{RGB}{100, 149, 237}
\definecolor{green-light-primary}{RGB}{240, 255, 240}    
\definecolor{green-light-secondary}{RGB}{220, 255, 220}
\definecolor{lightblue}{RGB}{222,235,247}
\usepackage{listings}
\usepackage{tcolorbox}
\usepackage[table]{xcolor}
\newtcolorbox{Mycolorbox}[2][]{
  arc=5mm,
  lower separated=false,
  fonttitle=\bfseries,
  colbacktitle=gray,
  coltitle=white,
  enhanced,
  attach boxed title to top left={xshift=0.5cm, yshift=-2mm},
  colframe=gray,
  colback=white,
  title=#1, #2,
  breakable
}

\usepackage{subcaption}
\usepackage[capitalize,noabbrev]{cleveref}
\title{Reasoning through Verifiable Forecast Actions: Consistency-Grounded RL for Financial LLMs}

\author{
  Jialin Chen\textsuperscript{1},
  Aosong Feng\textsuperscript{1},
  Harshit Verma\textsuperscript{1},
  Siyi Gu\textsuperscript{1},
  Haiwen Wang\textsuperscript{1},\\
  \textbf{Ali Maatouk\textsuperscript{1}, 
  Yifeng Gao\textsuperscript{2},
  Yixuan He\textsuperscript{3},
  Leandros Tassiulas\textsuperscript{1},
  Rex Ying\textsuperscript{1}}
  \\
  \textsuperscript{1}Yale University,
  \textsuperscript{2}University of Texas Rio Grande Valley,
  \textsuperscript{3}Arizona State University
}
\newcommand{\name}{\textbf{\textsc{Stock-R1}}\xspace}

\begin{document}

\maketitle

\begin{abstract}
Financial markets are characterized by extreme non-stationarity, low signal-to-noise ratios, and strong dependence on external information such as news, company fundamentals, and macroeconomic signals. Yet, existing approaches either abstract time-series into text or decouple forecasting from language-based reasoning, leading to a fundamental mismatch between qualitative reasoning and quantitative outcomes.
To address this, we introduce \name, a time-series–enhanced LLM that unifies stock forecasting and financial reasoning through a verifiable forecast action. 
Based on a tool-call design, the model first emits a forecast action, which is a structured and interpretable representation of its qualitative market outlook. It then invokes a time-series decoder conditioned on this action to generate distributional future trajectories, leading to more informed question answering and financial reasoning. We optimize the full pipeline with reinforcement learning, where rewards jointly reflect answer validity, forecast accuracy, and consistency between generated actions and observed time-series dynamics. In addition, rewards are reweighted by a sample-level uncertainty scalar, encouraging the model to accommodate varying uncertainty in market dynamics.
We evaluate \name on financial question answering and stock forecasting over a large-scale 10-year benchmark. Our method consistently outperforms time-series baselines and general-purpose LLMs, improving reasoning accuracy by 17.7\% (4B) and 25.9\% (8B). These findings demonstrate that structuring the forecast actions establishes a powerful synergy between language reasoning and temporal prediction, enabling LLMs to reason through verifiable, interpretable, and numerically grounded decisions.
\end{abstract}
\section{Introduction}
Many high-stakes decision-making problems require reasoning over evolving systems whose behavior cannot be explained by a single modality. Financial markets provide a particularly challenging instance of this setting: price movements are noisy, non-stationary, and tightly coupled with news, macroeconomic conditions and company fundamentals. Consequently, forecasting based only on past price trajectories remains difficult for both statistical models~\cite{taylor2017prophet,ariyo2014stock,box2015time} and recent deep time-series architectures~\cite{zeng2023transformers,liu2023itransformer,liu2022non,nie2022time}, as the predictive signals often extend beyond the observed numerical history.

\begin{wrapfigure}{r}{0.45\textwidth}\vspace{-0.3cm}
    \centering
\includegraphics[width=\linewidth]{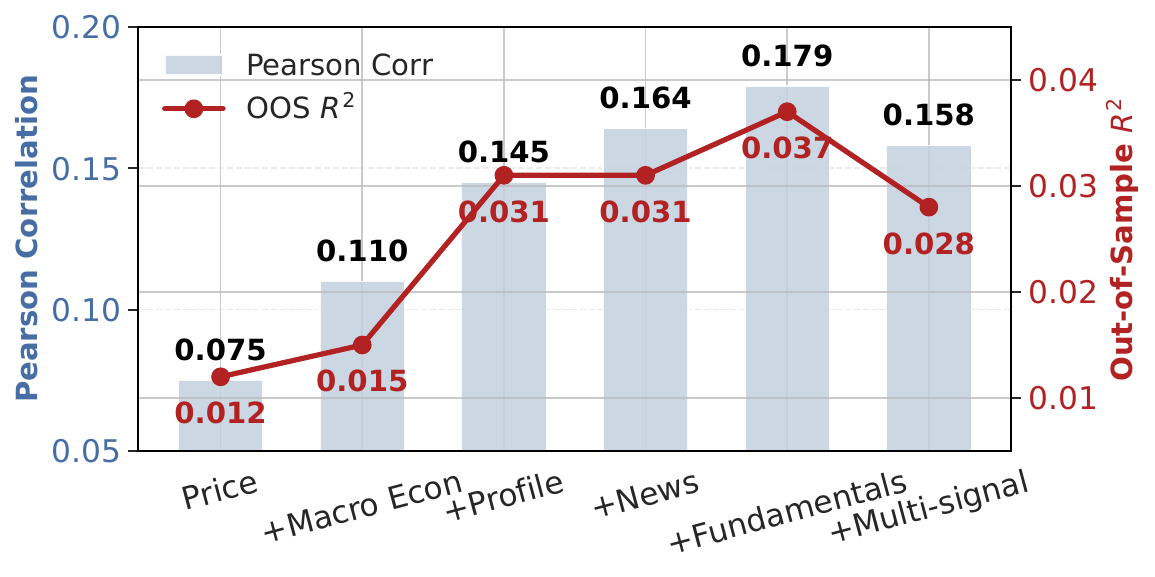}\vspace{-0.3cm}
    \caption{Pearson correlation and out-of-sample (OOS) $R^2$ for stock return prediction under different context settings.}
    \label{fig:info_rate}\vspace{-0.5cm}
\end{wrapfigure}

To examine whether external context provides measurable predictive information, we conduct a simple non-parametric diagnostic experiment. A k-nearest-neighbor regressor~\cite{cunningham2021k} predicts future cumulative returns by averaging outcomes from historically similar market states, where similarity is computed either from price history alone or from price history together with other textual contexts. Figure~\ref{fig:info_rate} shows that adding company-specific information, including business profiles, news, and fundamentals, consistently improves Pearson correlation between predicted and ground-truth returns as well as out-of-sample (OOS) $R^2$~\cite{campbell2008predicting}. This result shows that textual market context carries non-trivial predictive signals beyond historical price dynamics, highlighting its value for financial forecasting.

These observations motivate multimodal financial modeling, but existing approaches still lack a principled mechanism for coupling numerical forecasting with language-based reasoning. One line of work augments time-series models with news or fundamentals as auxiliary covariates~\cite{li2024causalstock,xu2018stock,li2023pen}, improving predictive performance but offering limited capacity for open-ended reasoning. Another line builds financial LLMs~\cite{wu2023bloomberggpt,liu2025fin,qian2025fino1,zhu2025dianjin,xiao2025trading} that can synthesize market narratives, yet they remain limited in representing dense numerical sequences and performing precise quantitative reasoning~\cite{wallace2019nlp, tan2024language,chen2025mtbench}. Recent RL-based LLM approaches for time-series forecasting further improve temporal reasoning by training LLMs to generate reasoning traces or direct forecasts~\cite{guan2025timeomni,luo2025time}. However, they still collapse semantic reasoning and numerical trajectory modeling into a single token-prediction objective, leaving their interaction weakly structured and difficult to verify. This suggests a natural question: \textit{can we design an explicit, verifiable interface through which an LLM contributes contextual reasoning about market dynamics, while delegating trajectory-level numerical generation to a dedicated time-series decoder?} 

However, this raises a central cross-modal alignment challenge. The LLM reasons over discrete semantic abstractions, whereas the time-series decoder generates continuous temporal trajectories. To enable effective synergy between the two modalities, the LLM must be able to use its contextual understanding to guide the decoder toward informed numerical generation. Conversely, the generated trajectory should remain consistent with the context, reducing mismatches between reasoning and forecasting. More broadly, this reflects a common challenge in cross-modal reasoning: aligning semantic reasoning with modality-specific generation without introducing hallucinated outputs.

To address this gap, we introduce \name, a financial LLM that couples language reasoning and numerical forecasting. The core design is a \emph{structured forecast action}, a field-factorized discrete profile describing key temporal attributes, such as direction, drawdown, and turning-point behavior, produced by the knowledgeable policy model. This action serves as an executable signal for the time-series decoder, allowing the LLM’s semantic assessment to guide numerical trajectory generation. At the same time, the action can be evaluated against modality-derived evidence at the field level, providing a verifiable bridge between qualitative reasoning and quantitative forecasting.

Training proceeds in two stages. We first perform supervised fine-tuning (SFT) using modality-derived forecast profiles to warm up action generation and downstream reasoning. We then apply a GRPO-based~\cite{shao2024deepseekmath} reinforcement learning stage, where the policy is optimized with a reward that combines answer validity, forecast accuracy, and action-level consistency between discrete fields and observed time-series behavior. To improve robustness under noisy market regimes, we incorporate \emph{uncertainty-aware reweighting} into the RL objective, which down-weights samples of high-uncertainty during policy updates. Together with the structured action space, this design enables more precise credit assignment than unstructured free-from rationales. Empirically, \name achieves state-of-the-art performance across price movement forecasting and financial QA on a newly curated large-scale benchmark across 10-year S\&P 500 stocks. These results suggest that verifiable forecast actions provide an effective interface for coupling language reasoning with temporal prediction. 

 Our contributions are threefold:
(1) We introduce forecast actions as an executable interface for effectively coupling language-based reasoning with trajectory-level time-series forecasting.
(2) We propose a consistency-grounded reinforcement learning objective with uncertainty-aware reweighting, which aligns predictive commitments, generated trajectories, and final answers under noisy market dynamics.
(3) We curate a large-scale financial benchmark with aligned temporal signals and textual context, together with an evaluation suite for forecasting, financial QA, and multimodal reasoning.




\section{Related Work}
\xhdr{General Time-Series Forecasting} The field of time-series forecasting has evolved from statistical baselines~\cite{ariyo2014stock, taylor2018forecasting} to sophisticated deep learning architectures. Transformer-based models have become the dominant paradigm, aimed at capturing long-range dependencies~\cite{zhou2021informer, wu2021autoformer, nie2022time, liu2023itransformer, feng2024efficient, Fedformer, kim2025comprehensive}.
With the rise of large language models, recent works have attempted to use pre-trained LLMs for time series~\cite{jin2023time, chang2023llm4ts,hu2025context,liu2024autotimes}, which typically align time-series embeddings with the text space of frozen LLMs. This line of research has recently progressed toward time-series foundation models~\cite{rasul2023lag,goswami2024moment,shi2024time,liu2024timer,woo2024unified,ansari2024chronos}, emphasizing large-scale pretraining across diverse datasets, stronger generalization, and transferability across tasks and domains. However, these methods primarily focus on forecasting benchmarks where patterns are relatively stationary, nor do they typically integrate auxiliary multimodal context to explain their predictions.

\xhdr{Stock Prediction and Quantitative Modeling} Financial forecasting presents unique challenges due to the stochastic nature of market dynamics. Previous approaches~\cite{ariyo2014stock, feng2019temporal, shi2025kronos,lee2025mitigating,duan2022factorvae} focus on extracting patterns from historical stock data, implicitly assuming that past prices contain sufficient information for future prediction, which contradicts the Efficient Market Hypothesis~\cite{malkiel1989efficient}. Recent multimodal approaches~\cite{kim2019hats,xu2018stock,li2024causalstock,wang2024stocktime} augment time-series models with external information such as textual news or relational graphs to capture cross-asset and event-driven dependencies. However, they typically treat auxiliary modalities as feature-level enhancements for forecasting or prediction. In contrast, our method is designed to natively handle reasoning-centric financial tasks, enabling coherent forecasting and answer generation.

\xhdr{Financial LLMs and Trading Agents} Financial LLMs have attracted growing interest recently~\cite{wu2023bloomberggpt, wang2023fingpt,liu2025fin, qian2025fino1,zhu2025dianjin}. While excelling in knowledge-intensive tasks, they struggle with forward-looking trend prediction. In parallel, research has investigated agent-based and RL-driven frameworks~\cite{sun2026trade,xiao2024tradingagents,xiao2025trading}, which focus on portfolio or risk optimization and interactive market play, but do not natively support fine-grained forecasting or reasoning for future market in context-rich QA settings. Our work explores an orthogonal and complementary direction, emphasizing grounded forecasting to support complex financial question answering.
\section{Methodology}
\begin{figure}[t]
    \centering
    \includegraphics[width=\linewidth]{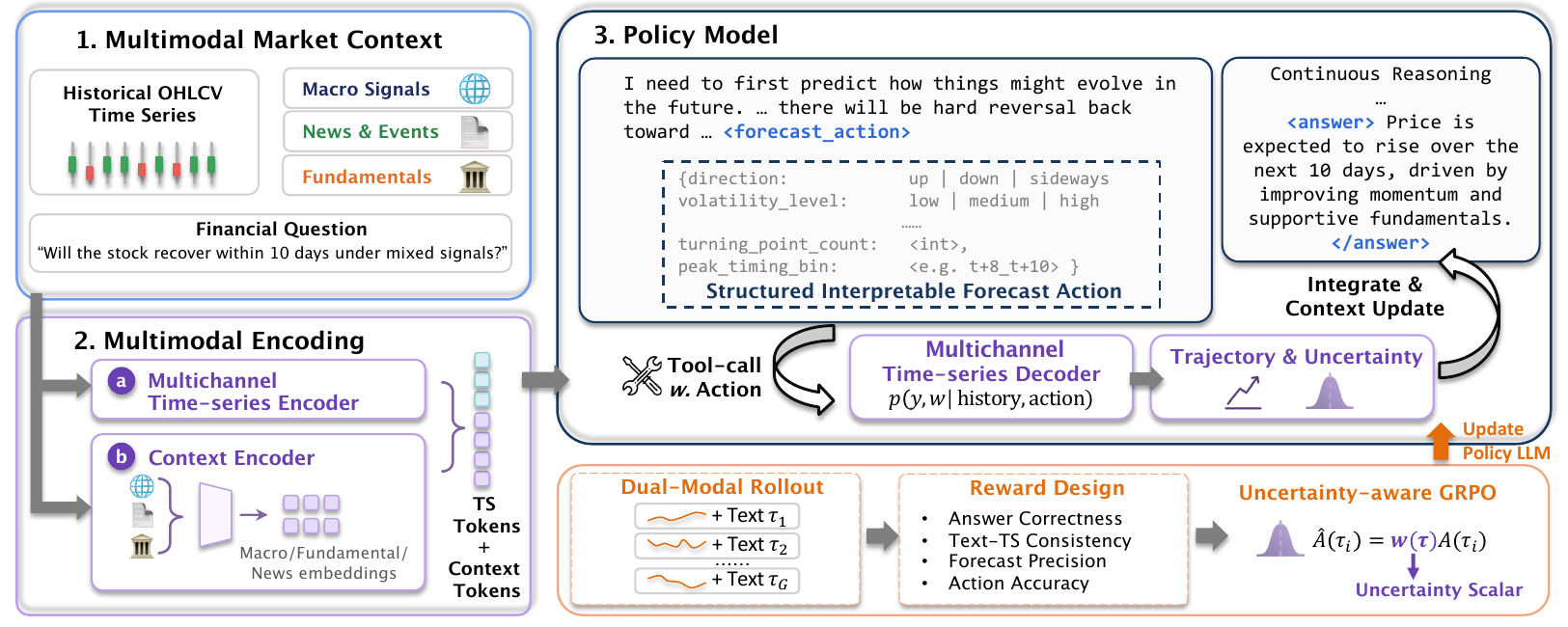}
\caption{
Overview of \name. The model encodes multimodal market context into unified tokens, then uses a policy LLM to generate a structured forecast action that conditions a multichannel time-series decoder. The predicted trajectory and uncertainty are returned to the LLM for grounded financial reasoning, with the whole policy optimized via uncertainty-aware dual-modal rollout.
}
    \label{fig:overall}\vspace{-0.3cm}
\end{figure}

As illustrated in Figure~\ref{fig:overall}, \name operates as an action-conditioned tool-use pipeline for financial reasoning. Given multimodal stock inputs (Sec.~\ref{sec:reasoning_data}), we first pre-train a multi-channel time-series encoder that generates future trajectories and uncertainty estimates (Sec.~\ref{sec:ts_encoder_pretrain}). The policy LLM then emits a structured forecast action that invokes the time-series decoder (Sec.~\ref{sec:forecast_action}), whose trajectory output is returned as numerical evidence for final answer generation. The pipeline is initialized with joint supervised fine-tuning (Sec.~\ref{sec:joint_sft}) and further optimized through consistency-grounded reinforcement learning with uncertainty-aware updates (Sec.~\ref{sec:rl}).

\subsection{Multimodal Time-Series Data Formulation}\label{sec:reasoning_data}
The fundamental challenge in financial time-series forecasting stems from market non-stationarity. Raw series often exhibit shifting statistical patterns over time, making patterns learned from historical observations unreliable under future market conditions and substantially weakening model generalization. To mitigate this, we reframe the pre-training objective from deterministic point forecasting to probabilistic density estimation~\cite{lo1988stock}. Given a multichannel input representation, $\mathbf{X}_t=[O_{t-L:t},H_{t-L:t},L_{t-L:t},C_{t-L:t},V_{t-L:t}]\in\mathbb{R}^{5\times L}$, encompassing the Open, High, Low, Close, and Volume (OHLCV) over a lookback window $L$, we transform these raw signals into log-return formats, defined formally as $r_t^{\mathrm{N}}=\ln(\frac{\mathrm{N}_t}{C_{t-1}})$ for $\mathrm{N}\in\{O,H, L,C\}$ and $r_t^{V}=\ln(\frac{V_t+1}{V_{t-1}+1})$, which effectively converts the non-stationary sequence into a quasi-stationary signal suitable for efficient representation learning~\cite{fama1970efficient}.

To complement these numerical signals with semantic context, we align the time-series data with a structured modality $\mathcal{T}_t$. This includes: (1) \textbf{Macroeconomic Data}: Key indicators such as US Treasury yields (\eg 2Y, 10Y spreads) and inflation metrics that drive broad market sentiment ; (2) \textbf{Company Fundamentals}: Quarterly financial snapshots including revenue, operating margins, and leverage ratios (\eg debt-to-equity); and (3) \textbf{Business \& News Context}: A summary of the company's core operations alongside recent headlines and sentiment analysis. These heterogeneous data sources are serialized into a unified textual prompt, serving as the grounding context for the model's reasoning. Data collection details and examples are provided in Appendix~\ref{Appx:data_details}.


\subsection{Multi-Signal Time Series Encoder Pre-training}\label{sec:ts_encoder_pretrain}
\xhdr{Time Series Encoder Architecture} We employ a causal attention-based transformer that uniformly encodes the multichannel input $\mathbf{X}_t \in \mathbb{R}^{5 \times L}$, enabling the model to learn joint representations across both temporal and variable dimensions. To rigorously preserve temporal consistency within this flattened space, we implement Rotary Positional Embeddings (RoPE)~\cite{su2024roformer} where positional indices are assigned strictly according to the timestamp $t$. The core Multi-variate Time Attention mechanism refines standard self-attention by injecting a learnable binary bias $\mathbf{B}$ into the attention scores, which explicitly differentiates between intra-variable dependencies (where variable indices $v_i = v_j$) and inter-variable cross-correlations (where $v_i \neq v_j$). This is complemented by a causal block-diagonal mask $\mathbf{M}$ that enforces strict autoregressive constraints, ensuring that predictions at time $t$ depend solely on the history $\tau \leq t$ across all channels. More details are given in Appendix~\ref{appd:ts_encoder}.

\xhdr{Pretraining Objectives} We employ a reconstructive objective to train the time-series encoder, ensuring that the resulting flattened latent tokens $\mathbf{H}_\text{ts} \in \mathbb{R}^{K \times d_m}$, where $K$ represents the number of time-series tokens and $d_m$ is the LLM's model dimension. These latent tokens are used as time-series tokens and prepended to the textual tokens as part of the LLM input, allowing the policy model to jointly attend to numerical market dynamics and semantic market context. Traditional point-wise forecasting often fails in financial contexts because it ignores the inherent stochasticity of market dynamics. To address this, we project $\mathbf{H}_\text{ts}$ to model the conditional probability of future returns as dynamic Gaussian distributions $\mathcal{N}(\mu_{t+s}, \sigma^2_{t+s})$ for Open, Close, and Volume. For the intraday extremes (High and Low), which reflect the volatility boundaries, we employ quantile regression to estimate tail risks. The training objective $\mathcal{L}_{\text{TS}}$ combines the Negative Log-Likelihood (NLL) and the quantile loss $\rho_\tau$ for the volatility boundaries over the horizon $S$:
\begin{equation*}
\mathcal{L}_{\text{TS}} = \sum_{s=1}^S \sum_{i \in \{O,C,V\}} \lambda_i \left( \frac{(r^i_{t+s} - \mu^i_{t+s})^2}{2(\sigma^i_{t+s})^2} + \frac{1}{2}\ln\left((\sigma^i_{t+s})^2\right) \right) + \sum_{j \in \{H, L\}} \lambda_j \rho_{\tau_j}\left(r^j_{t+s} - q^{(j, \tau_j)}_{t+s}\right),
\end{equation*}
where the quantile loss function $\rho_\tau(u) = u(\tau - \mathbb{I}_{u < 0})$. Specifically, we target the 95th percentile for the High price and the 5th percentile for the Low (\ie $\tau_H=0.95;\tau_L=0.05$). This formulation not only provides a predictive mean but also allows us to explicitly quantify market uncertainty, defined as the aggregate variance $\mathcal{U} = \sum_{s,i} (\sigma^i_{t+s})^2$, which is further normalized to a bounded range to serve as a calibrated stability signal for downstream reinforcement learning.

\subsection{Structured Forecast Actions}\label{sec:forecast_action}

\xhdr{Single-Action Tool-Call Interface}
The pretrained time-series model generates future trajectories from historical signals. However, it does not utilize how market context should shape the forecast. We therefore formulate forecasting-enhanced reasoning as a single-action tool-call process, where the LLM emits a structured forecast action as the argument to a dedicated time-series decoder. This action serves as a semantic-to-numeric interface, converting the LLM's contextual market assessment into an executable control signal that modulates the decoder's forecast distribution.

\xhdr{Action Space Definition}
Let the forecast action consist of $K$ forecast-relevant fields, where each field describes one temporal attribute of the future trajectory, such as trend direction, end-point movement, turning-point behavior, or peak timing. For the $k$-th field, we define its value set as $\mathcal{A}_k$, which is finite and can be either categorical or ordinal. For example, a categorical field such as trend direction has $\mathcal{A}_k=\{\textit{up},\textit{flat},\textit{down}\}$, while an ordinal field such as volatility level width uses discretized bins $\mathcal{A}_k=\{1,\ldots,B_k\}$. The complete forecast action space is the Cartesian product over all fields $\mathcal{A}_{\mathrm{forecast}}=\mathcal{A}_1\times\cdots\times\mathcal{A}_K$. 

Given the multimodal context $s$ and user query $q$, the policy LLM $\pi_\theta$ generates a forecast action $a=(a_1,\ldots,a_K)\in \mathcal{A}_{\mathrm{forecast}}$ by $ a \sim \pi_\theta(\cdot\mid s, q)$.
This field-factorized construction makes the action interpretable, exposes fine-grained temporal commitments, and enables field-level supervision and reward computation that would be difficult to obtain from free-form natural-language rationales.

\xhdr{Action-conditioned Forecasting Tool}
The forecast action invokes the time-series decoder in a single tool-call interface (by generating  \texttt{<forecast\_action>}).  Given the encoded historical embedding $\mathbf{H}_{\mathrm{ts}}$ and a structured action $a$, we embed its categorical and ordinal fields, concatenate them into a unified action representation, and feed it into an MLP action encoder $e_\psi(\cdot)$ to obtain the continuous conditioning vector $\mathbf{z}_a$ for the decoder. The time-series decoder $\phi(\cdot)$ then generates future trajectories conditioned on both the historical time-series representation and the encoded action $\hat{y}_{t+1:t+S}\sim p_\phi(\cdot\mid \mathbf{H}_{\mathrm{ts}},e_\psi(a))$.
Thus, different forecast actions can induce different forecast distributions under the same historical context. After execution, the decoder output is serialized as the observation returned by \texttt{<forecast\_action>}, and the LLM completes the final answer conditioned on the original context, the structured action, and the numerical trajectory.

\xhdr{Why Forecast Actions}
The central challenge is bridging the LLM's semantic reasoning and the decoder's numerical generation while preserving cross-modal faithfulness. Forecast actions provide this missing interface by summarizing the LLM's market outlook into structured temporal attributes, which directly condition the decoder's generation. This intermediate action space renders the interaction (i) \textbf{controllable}, as distinct actions induce distinct predictive distributions over future trajectories; (ii) \textbf{verifiable}, as each field of the action can be evaluated against both the subsequently generated forecast and the realized outcome; and (iii) \textbf{trainable}, as reinforcement learning can assign credit to each attribute rather than solely at the entire prediction.

\subsection{Joint SFT with Context-Aware Forecasting Decoder}\label{sec:joint_sft}

To bootstrap supervised training in the financial domain, we distill cold-start trajectories from GPT-5 under the same tool-call format. Given a multimodal market context, the teacher generates financial questions covering diverse perspectives, constructs action-grounded reasoning, and then completes the reasoning and answers specific to the question. During supervised data construction, the target action is deterministically derived from the realized future trajectory, providing field-level supervision for the LLM's intermediate forecast decision. During inference, the LLM emits the structured action before invoking \texttt{<forecast\_action>}.  More details are given in Appendix~\ref{appd:cold_data}.

In this stage, the policy LLM is trained to generate a forecast rationale, emit valid and precise structured forecast action $a$, consume the decoder trajectory returned by \texttt{<forecast\_action>}, and produce the final reasoning trace and answer. The SFT objective contains two coupled terms. First, $\mathcal{L}_{\text{TS}}$ (in Sec.~\ref{sec:ts_encoder_pretrain}) trains the encoder and decoder to model future trajectories using the distribution-aware loss, now conditioned on the forecast action when available. Second, $\mathcal{L}_{\text{SFT}}$ trains the LLM with a next-token prediction objective to follow the reasoning-action-tool-answer format. The overall objective is thereby $\mathcal{L}_{\text{total}}=\mathcal{L}_{\text{SFT}}+\lambda_{\text{ts}}\mathcal{L}_{\text{TS}}$.

\subsection{Reinforcement Learning for Grounded Reasoning}\label{sec:rl}
After SFT, the model can follow the tool-call format, but supervised learning alone does not guarantee that the generated action, decoder trajectory, and final answer remain mutually consistent. We therefore introduce an uncertainty-aware reinforcement learning (RL) stage over the full dual-modal rollout. For each state $s=(\mathbf{X}_t,\mathcal{T}_t,q)$, the policy first samples a textual forecast action $\hat{a}$, the time-series decoder then generates a numerical trajectory $\hat{y}_{t+1:t+S}\sim p_\phi(\cdot\mid \mathbf{H}_{\mathrm{ts}},e_\psi(\hat{a}))$, and the LLM finally produces an answer $\hat{z}$ conditioned on the original context, action, and the time-series rollout. This dual-modal rollout explicitly couples the LLM's reasoning capability with the action-conditioned decoder: a poor action can degrade the forecast, and an inconsistent forecast can reduce the reward assigned to the reasoning trace. At the same time, financial labels are intrinsically noisy under high-volatility regimes; hence, the RL update should account for the calibrated uncertainty $\mathcal{U}_q$ estimated by the distribution-aware time-series module in Sec.~\ref{sec:ts_encoder_pretrain}.

\xhdr{Reward Design}
The reward evaluates both modalities. We define the total reward as
\begin{equation*}
R = R_{\mathrm{ans}} + \alpha R_{\mathrm{act}} + \beta R_{\mathrm{prec}} + \gamma R_{\mathrm{cons}},
\end{equation*}
where $\alpha$, $\beta$, and $\gamma$ are weighting coefficients that balance the contributions of each reward. Each component targets a distinct failure mode:
\begin{itemize}[leftmargin=0.5cm, nosep]
    \item \textbf{Answer Correctness ($R_{\mathrm{ans}}$)} evaluates the final answer according to the task type. For categorical or decision-oriented queries, we use exact matching over the target class; for open numerical questions, correctness is computed with tolerance-aware scoring.
    \item \textbf{Action Accuracy ($R_{\mathrm{act}}$)} evaluates whether the generated forecast action $\hat{a}$ matches the ground-truth action $a^\star=g(y_{t+1:t+S})$ derived from the realized future trajectory. We compare the two actions field by field: categorical fields are scored by exact matching, while ordinal fields receive distance-aware partial credit.
    \item \textbf{Forecast Precision ($R_{\mathrm{prec}}$)} measures the accuracy of the decoder-generated trajectory against the realized time series. We use scale-normalized errors for price and volume and direction-aware metrics for return movement, so the reward remains comparable across assets.
    \item \textbf{Action-Trajectory Consistency ($R_{\mathrm{cons}}$)} verifies whether the structured action is consistent with the time-series trajectory produced by the decoder. We parse the generated trajectory into a profile $\tilde{a}_h=g(\hat{y}_{t+1:t+S})$ and compare it with the policy action $\hat{a}_h$ field by field. Categorical fields are scored by exact agreement, while ordinal fields receive distance-sensitive similarity.
\end{itemize}
The consistency term is central to the proposed alignment mechanism. It prevents the LLM from emitting a bullish action while the decoder returns a bearish trajectory, or from claiming low volatility while generating a wide forecast range. Because the reward depends jointly on the policy-selected action and the decoder-conditioned forecast, it provides credit assignment across the interface between the LLM parameters $\theta$ and the decoder parameters $\phi$.

\xhdr{Uncertainty-Aware Consistency-Grounded GRPO}
For each query, we sample a group of dual-modal rollouts $\{o_i\}_{i=1}^{G}$, where each rollout contains the generated action, numerical trajectory, and final answer. The group-relative advantage is computed from the total reward above, allowing the model to compare alternative actions and answers under the same market state. To avoid overfitting noisy credit assignment in unstable market regimes, we optimize the policy with a GRPO-style objective~\cite{shao2024deepseekmath} by further reweighting the learning signal with the normalized forecast uncertainty $\mathcal{U}_q$ predicted by the pre-trained time-series encoder, which is defined as:
\begin{equation*}
\mathcal{J}_{\text{u-GRPO}}(\theta)= \mathbb{E}_{q,\{o_i\}}\boxed{w(\mathcal{U}_q)}
\frac{1}{G} \sum_{i=1}^G \frac{1}{|o_i|} \sum_{t=1}^{|o_i|}
\left\{
\min \left[ r_t(\theta) \hat{A}_{i,t},\text{clip}(r_t(\theta)) \hat{A}_{i,t} \right]
- \beta \mathbb{D}_{KL} [\pi_\theta || \pi_{ref}]
\right\}.
\end{equation*}
We normalize the sample uncertainty weight $w(\mathcal{U}_q)$ using a threshold-based exponential decay function (see Appendix~\ref{appx:rl_details} for the detailed formulation), which effectively down-weights contributions from high-volatility samples. This mechanism ensures that policy updates are driven primarily by identifiable and stable market regimes, while maintaining conservative updates during highly stochastic periods. Integrating the estimated volatility effectively forces the policy to discover consistent and reliable structural patterns where ground-truth outcomes can be logically inferred from the context.

\section{Experiments}\label{sec:experiments}
We evaluate \name on complex financial QA (Sec.~\ref{sec:QA}), time-series forecasting (Sec.~\ref{sec:forecasting}), followed by ablation studies on time-series input and analysis on the quality of forecast actions (Sec.~\ref{sec:ablation}). We further provide a model scalability analysis in Appendix~\ref{appd:scale_ablation} and an investment simulation in Appendix~\ref{sec:simulation} to evaluate the quality of the model's predictive buy/sell signals.

\subsection{Reasoning-based Financial QA}\label{sec:QA}
\xhdr{Experimental Setting} We evaluate predictive and financial question answering across five task categories, which are designed to assess both numerical foresight and the ability to integrate heterogeneous financial signals. We compare \name against strong language-model baselines, including general-purpose LLMs (Llama3~\cite{dubey2024llama}, Qwen3~\cite{yang2025qwen3}, GPT-5, OpenAI-o3) and domain-adapted financial reasoning models (Fin-o1~\cite{qian2025fino1}, Fin-R1~\cite{liu2025fin}, DianJin-R1~\cite{zhu2025dianjin}). \name is instantiated on top of Qwen3-4B~\cite{yang2025qwen3} and Qwen3-8B as the base language models. All models are evaluated using task-specific accuracy. Experiment configurations are provided in Appendix~\ref{appd:hyper_setting}.
\begin{wraptable}{r}{0.65\textwidth}
\centering
\setlength{\tabcolsep}{0.8pt}
\caption{Performance comparison across financial QA tasks. Best performance is in \textbf{bold}, while the second-best is \underline{underlined}.}
\label{tab:QA}
\resizebox{0.65\columnwidth}{!}{%
\begin{tabular}{lcccccc}
\toprule
\textbf{Model} & \textbf{Forecast } & \textbf{Event}& \textbf{Fundamental} & \textbf{News} & \textbf{Multi-signal}  & \multirow{2}{*}{\textbf{Avg.}} \\
& \textbf{QA} & \textbf{Detection}&\textbf{QA} & \textbf{Analysis} & \textbf{Reasoning} & \\
\midrule
\multicolumn{7}{l}{\textit{\textbf{General-purpose LLM}}} \\
Llama3-3B    & 2.53  & 48.87 & 45.68 & 42.72 & 56.54 & 39.27 \\
Llama3-8B    & 3.21  & 44.10 & 47.13 & 51.77 & 54.81 & 40.20 \\
Qwen3-4B & 7.92  & 65.58 & 48.94 & 35.11 & 57.82 & 43.07 \\
Qwen3-8B & 6.23  & 39.42 & 47.06 & 44.79 & 52.69 & 38.04 \\
Qwen3-30B    & 10.34 & 55.35 & 45.40 & 51.41 & 63.39 & 45.18 \\
Llama3-70B	&5.35	&53.25	&59.34	&55.28	&60.85&	46.81	\\	
DeepSeek-V3.2 &3.37	&56.90	&44.58&	42.00	&60.95	&41.56\\
GPT-5        & 8.16  & 51.08 & 52.69 & 60.49 & 61.24 & 46.73 \\
OpenAI-o3 &2.78	&53.43&	57.76&	57.89&	67.07	&47.79\\
\midrule
\multicolumn{7}{l}{\textit{\textbf{Financial LLM}}} \\
Fin-R1-8B  & 40.28 & 49.35 & 56.41 & 60.48 & 60.28 & 53.36\\
Fin-o1-14B	&32.86	&31.89	&41.27	&53.53	&\textbf{69.12}&45.73 \\
DianJin-R1-32B&38.42 & 63.71&52.47&\underline{62.40} & 62.58 &55.92 \\
\rowcolor{lightblue!60}\name-4B &\underline{51.00} & \textbf{68.52}&\underline{59.70}&61.83&63.02&\underline{60.81}\\
\rowcolor{lightblue!60}\name-8B &\textbf{56.78} &\underline{67.49} & \textbf{61.08}&\textbf{67.16}&\underline{67.45}&\textbf{63.99}\\
\bottomrule
\end{tabular}}\vspace{-0.3cm}
\end{wraptable}

\textbf{Results.} Table~\ref{tab:QA} shows that \name consistently achieves the strongest overall performance across predictive and reasoning-centric financial QA tasks. The most pronounced gain appears in Forecast QA, where general-purpose LLMs perform poorly despite their large scale, suggesting that textual reasoning alone is insufficient for precise forward-looking numerical prediction. In contrast, \name substantially improves this category by grounding the reasoning process in trajectory-level time-series forecasts. Beyond pure forecasting, \name also performs strongly on Event Detection, Fundamental QA, News Analysis, and Multi-signal Reasoning, indicating that the proposed action-conditioned forecasting interface does not merely improve numerical prediction but also supports broader context-aware financial reasoning. Compared with domain-adapted financial LLMs, \name achieves higher average accuracy at both 4B and 8B scales, demonstrating that explicitly coupling language reasoning with dedicated temporal generation is more effective than relying on financial instruction tuning alone. Overall, these results suggest that verifiable forecast actions provide a practical mechanism for aligning market-context interpretation with numerical forecasting evidence. See Appendix~\ref{appd:case_study} for case studies in QA tasks.

\subsection{Time Series Forecasting}\label{sec:forecasting}
\xhdr{Baselines} We conduct a comprehensive evaluation of \name on the time-series forecast task against three distinct categories of baselines: (1) Time Series Foundation Models evaluated in a zero-shot setting (\eg Moirai~\cite{woo2024unified}, Chronos~\cite{ansari2024chronos}); (2) Time Series Models trained from scratch. Specifically, we include general forecasters, ARIMA and Prophet~\cite{box2015time,taylor2017prophet}, and re-pretrain the stock-specific model Kronos~\cite{shi2025kronos}. We also include state-of-the-art baselines such as AutoTimes~\cite{liu2024autotimes}, Time-LLM~\cite{jin2023time}, and FSCA~\cite{hu2025context}. These methods leverage LLMs to incorporate textual summaries, but do not naturally support the ingestion of lengthy, temporally-aligned multimodal context, often limiting their ability to model high-frequency market dynamics; and (3) Multimodal Large Language Models (\eg GPT-5, Qwen3~\cite{yang2025qwen3}) that understand time series in text space. We report separate results for our model in Unimodal \textbf{(U)} and Multimodal \textbf{(M)} settings to quantify the gain from textual grounding. Performance is measured using Mean Absolute Percentage Error (MAPE) and Root Mean Square Error (rMSE) for numerical precision (Price, Volume, Returns), Directional Accuracy (Acc.) for trend prediction horizons of next 3, 5, and 10 days, and Mean Absolute Error (MAE) for volatility estimation. More experimental details are given in Appendix~\ref{appx:baselines}.

\begin{table*}[t]
\centering\vspace{-0.1cm}
\caption{Holistic Forecasting Performance Comparison. We evaluate \name against Zero-shot Foundation Models, Supervised Time Series Models, and Multimodal LLMs. \textbf{(U)} denotes Unimodal input (Time Series only), and \textbf{(M)} denotes Multimodal input (Time Series + Text). The best performance is in \textbf{bold}, while the second-best performance is \underline{underlined}.}
\vspace{-0.2cm}
\label{tab:forecasting_main}
\setlength{\tabcolsep}{2pt}
\resizebox{\textwidth}{!}{%
\begin{tabular}{clccccccccc}
\toprule
& Model 
& Price $\downarrow$ 
& Price $\downarrow$ 
& Volume $\downarrow$ 
& Volume $\downarrow$ 
& Return $\downarrow$ 
& 3-day $\uparrow$ 
& 5-day $\uparrow$ 
& 10-day $\uparrow$ 
& Volatility $\downarrow$ \\
&  
& (MAPE) 
& (rMSE) 
& (MAPE) 
& (rMSE) 
& (rMSE) 
& Acc. (\%) 
& Acc. (\%) 
& Acc. (\%) 
& (MAE$\times 10^{-3}$) \\
\midrule

\multirow{5}{*}{
\begin{tabular}{c}
Time Series \\
Foundation Models \\
(\textbf{Zero-shot})
\end{tabular}}
& Timer-XL 
& 0.946 & 1.460 & 6.758 & 1.392 & 0.955 & 44.7 & 43.8 & 40.3 & 7.212 \\
& Moirai 
& {0.031} & 1.395 & 7.293 & \underline{1.367} & 0.048 & 53.2 & 52.8 & 54.5 & 2.709 \\
& MOMENT 
& 0.076 & 1.484 & 6.150 & 1.409 & 0.104 & 49.2 & 48.5 & 46.6 & 2.937 \\
& Chronos-T5 
& {0.031} & 1.411 & 7.097 & 1.376 & 0.051 & 50.5 & 50.1 & 49.6 & 6.453 \\
& Time-MoE	
& 0.035 & 1.437 & 5.759 & \textbf{1.358} & 0.052 & 48.5 & 48.6 & 48.4 & 3.008 \\

\midrule
\multirow{12}{*}{
\begin{tabular}{c}
Time Series Models \\
(\textbf{Train-from-scratch})
\end{tabular}}
& ARIMA 
& 0.035 & 1.405 & 6.134 & 1.415 & 0.061 & 53.4 & 53.0 & 50.0 & -- \\
& Prophet 
& 0.058 & 1.416 & 6.128 & 1.377 & 0.083 & 47.3 & 45.9 & 49.9 & 2.966 \\
& DLinear 
& 0.038 & 1.433 & 6.019 & 1.388 & 0.055 & 51.8 & 51.7 & 50.8 & 3.015 \\
& Mamba 
& 0.037 & 1.448 & 6.345 & 1.464 & 0.053 & 52.3 & 53.0 & 53.3 & 3.146 \\
& TimeMixer 
& {0.031} & 1.401 & 6.087 & 1.403 & 0.049 & 53.8 & 54.6 & 55.2 & 2.946 \\
& TimesNet 
& {0.031} & 1.423 & 6.037 & 1.387 & 0.046 & 52.2 & 50.3 & 53.5 & 2.898 \\
& PatchTST 
& 0.037 & 1.409 & 5.991 & 1.395 & 0.055 & 52.9 & 51.1 & 51.7 & 14.124 \\
& NSTransformer 
& {0.031} & 1.391 & 6.074 & 1.399 & 0.048 & 53.3 & 51.1 & 53.2 & 2.928 \\
& iTransformer 
& 0.032 & 1.415 & 6.098 & 1.399 & 0.046 & 54.4 & 53.8 & 53.3 & 2.967 \\
& AutoTimes 
& 0.042 & 1.444 & 5.938 & 1.384 & 0.059 & 52.9 & 51.4 & 49.4 & {2.634} \\
& Time-LLM 
& 0.049 & 1.450 & 6.108 & 1.411 & 0.070 & 50.2 & 51.4 & 46.9 & 3.383 \\
& Kronos 
& 0.036 & 1.371 & 5.725 & 1.385 & 0.051 & 50.8 & 52.5 & 54.2 & {2.634} \\
& FSCA 
& \textbf{0.029} & 1.365 & 5.983 & 1.394 & 0.042 & 54.9 & 53.2 & 53.5 & 2.835 \\

\rowcolor{lightblue!60}\cellcolor{white}
& \name (U) 
& \textbf{0.029} & 1.339 & \textbf{5.316} & 1.384 & \underline{0.041} & 55.1 & 55.3 & 55.0 & \textbf{2.537} \\

\midrule\midrule
\multirow{6}{*}{
\begin{tabular}{c}
Language Models \\
(\textbf{Multimodal Input})
\end{tabular}}
& Qwen3-4B 
& 0.036 & {1.320} & 7.945 & 1.422 & 0.050 & \underline{56.4} & 55.6 & 58.6 & 2.778 \\
& Qwen3-8B 
& 0.044 & \textbf{1.303} & 8.373 & 1.417 & 0.060 & 55.6 & \underline{55.9} & \underline{59.2} & 2.682 \\
& Llama3-3B 
& 0.039 & 1.328 & -- & -- & 0.057 & 50.5 & 52.3 & 53.1 & 2.884 \\
& Llama3-8B 
& 0.035 & 1.324 & \underline{5.393} & 1.418 & 0.049 & 55.1 & 55.7 & 57.8 & 2.758 \\
& GPT-5 
& {0.032} & 1.334 & 8.409 & 1.411 & \underline{0.046} & 55.2 & 54.6 & 56.6 & 2.903 \\

\rowcolor{lightblue!60}\cellcolor{white}
& \name (M) 
& {0.031} & \underline{1.319} & 5.872 & 1.380 & \textbf{0.040} & \textbf{57.6} & \textbf{58.8} & \textbf{59.6} & \underline{2.602} \\
\bottomrule
\end{tabular}}
\vspace{-0.5cm}
\end{table*}

\xhdr{Results Analysis} As shown in Table~\ref{tab:forecasting_main}, among pure time-series approaches, \name \textbf{(U)} achieves state-of-the-art or highly competitive performance across numerical precision, return prediction, and volatility estimation. Notably, it obtains the best volatility MAE, suggesting that the distribution-aware pretraining objective better captures market risk dynamics than standard point-forecasting losses. Comparing \name \textbf{(U)} and \name \textbf{(M)}, multimodal context brings the clearest gains in directional accuracy across 3-, 5-, and 10-day horizons, while numerical precision remains broadly comparable. This pattern is consistent with the role of textual signals in financial markets: news, fundamentals, and macro context often provide directional cues about future movement, but do not necessarily determine exact price levels. Overall, \name \textbf{(M)} outperforms general-purpose LLMs by combining semantic market context with a dedicated time-series forecasting module, yielding stronger directional prediction without sacrificing trajectory-level numerical fidelity.

\subsection{Ablation Studies and Analysis}\label{sec:ablation}
\begin{figure}[ht]
    \centering
    \vspace{-0.3cm}
    \begin{subfigure}[t]{0.5\textwidth}
        \centering
        \includegraphics[width=\linewidth]{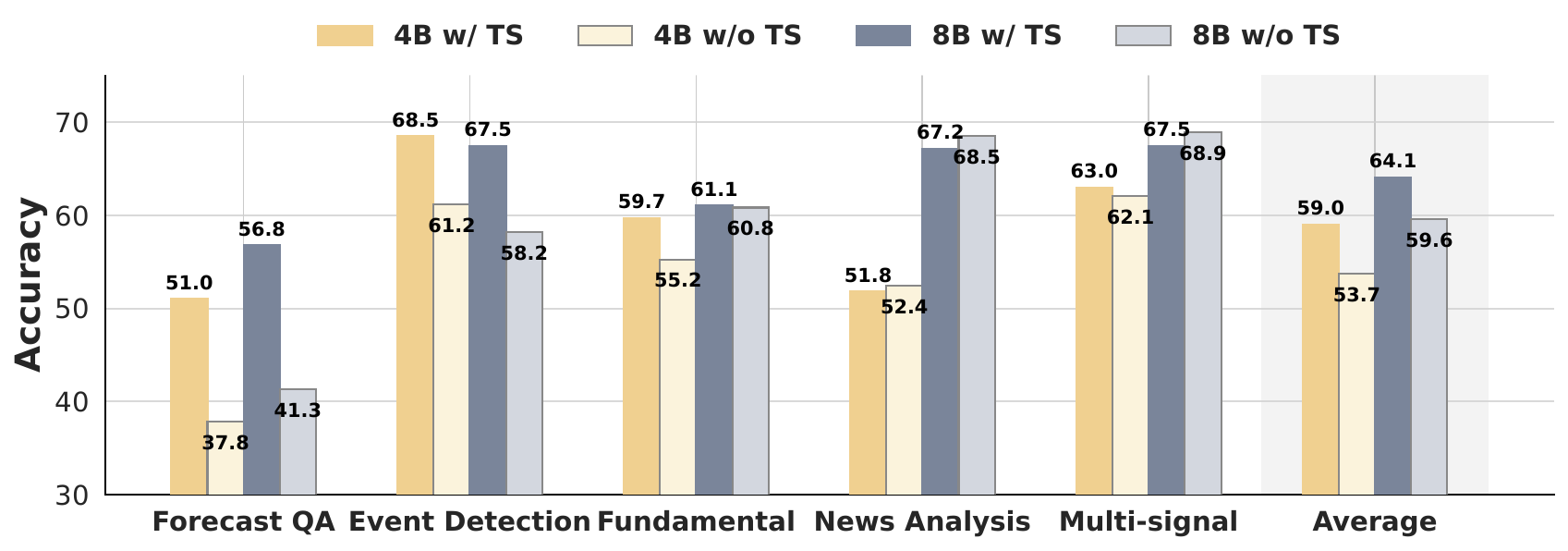}
        \caption{Performance comparison across QA tasks \textit{w/} and \textit{w/o} time-series token.}     \label{fig:ts_token_ablation}
    \end{subfigure}
    \hfill
    \begin{subfigure}[t]{0.25\textwidth}
        \centering
        \includegraphics[width=\linewidth]{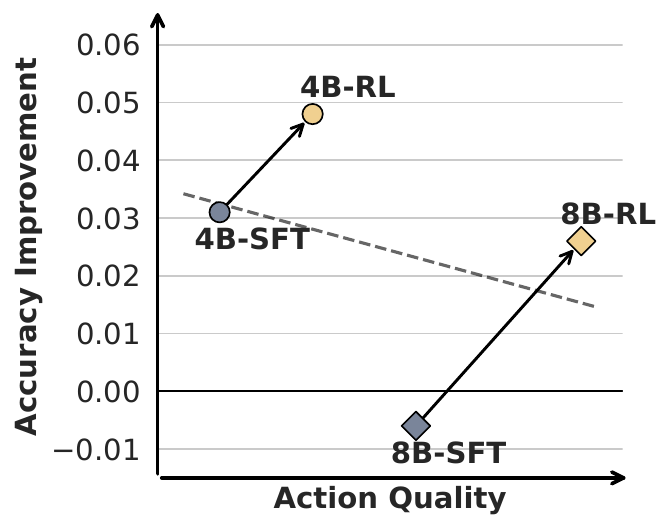}
        \caption{RL improves action quality beyond SFT}\label{fig:hint_quality_accuracy}
    \end{subfigure}
    \begin{subfigure}[t]{0.23\textwidth}
        \centering
        \includegraphics[width=\linewidth]{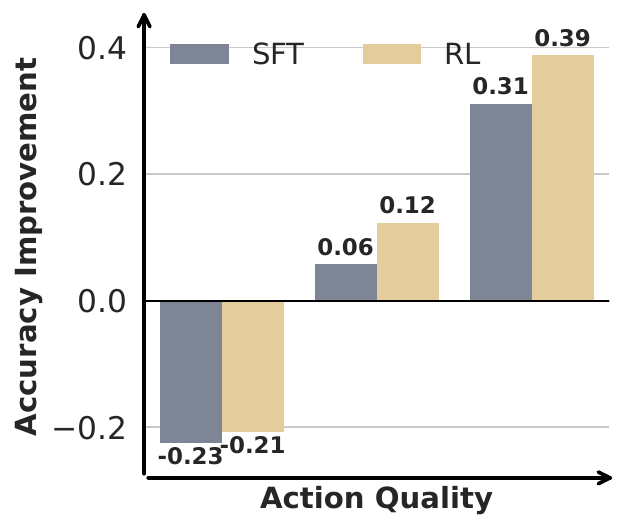}
        \caption{Better actions lead to higher accuracy}\label{fig:hint_bar}
    \end{subfigure}
    \vspace{-0.2cm}
    \caption{Ablation and Analysis of Time-Series Grounding and Action Quality}\label{fig:combine}
    \vspace{-0.5cm}
\end{figure}

\xhdr{Time-series Tokens Yields better Forecasting Performance} Figure~\ref{fig:ts_token_ablation} examines the impact of explicitly injecting time-series as soft tokens across different financial reasoning tasks. We observe consistent performance gains from time-series tokens at both 4B and 8B scales, indicating that the benefits are robust to model capacity. The largest relative improvements appear in Forecast QA, where access to explicit temporal dynamics is critical for anticipating future movements. However, without time-series tokens, models struggle to infer quantitative trends from textual context alone. For other types of questions, the gains are more moderate but still consistent. Overall, the average accuracy improves across all tasks and scales, demonstrating that time-series tokens serve as a general and task-agnostic inductive bias that enhances both predictive and reasoning-intensive financial QA.

\xhdr{RL Enhances Action Quality and Improves Directional Accuracy} As shown in Figure~\ref{fig:hint_quality_accuracy}, RL primarily acts as a mechanism for refining the model's forecast action space. We measure action quality by field-level accuracy, comparing each generated forecast-action field with the corresponding target field derived from the realized future trajectory. In particular, RL systematically shifts the model toward producing higher-quality and more internally consistent actions, suggesting that its core effect is to improve the precision and reliability of the intermediate decision interface. Moreover, Figure~\ref{fig:hint_bar} suggests a strong and consistent correlation between action quality and directional accuracy improvement, with Pearson correlation coefficients exceeding $0.65$ across settings. This indicates that better-aligned forecast actions translate into more accurate directional prediction, and the RL stage further amplifies this effect by enforcing qualitative--numerical consistency during training. Importantly, this relationship is monotonic and model-agnostic, suggesting that the structured action serves as an effective bottleneck that mediates between high-level reasoning and numerical prediction.

\begin{wrapfigure}{r}{0.4\textwidth}
\vspace{-0.5cm}
\centering
\begin{minipage}{0.4\textwidth}
    \centering
        \captionof{table}{Ablation study of RL variants and reward components.}
    \label{tab:ugrpo_ablation}
    \resizebox{0.9\linewidth}{!}{
    \begin{tabular}{lc}
    \toprule
    \textbf{Method} & \textbf{Avg. QA Acc.} \\
    \midrule
    SFT \textit{w/} thinking & $57.3 \pm 0.8$ \\
    \quad + GRPO & $66.0 \pm 1.1$ \\
    \quad + u-GRPO & $\mathbf{67.5 \pm 0.6}$ \\
    \midrule
    \textit{w/o} $R_{\text{act}}$ (Actions) & $63.2 \pm 0.8$ \\
    \textit{w/o} $R_{\text{prec}}$ (Forecast)& $66.4 \pm 1.1$ \\
    \bottomrule
    \end{tabular}
    }
    \par\vspace{0.2cm}
    \includegraphics[width=\linewidth]{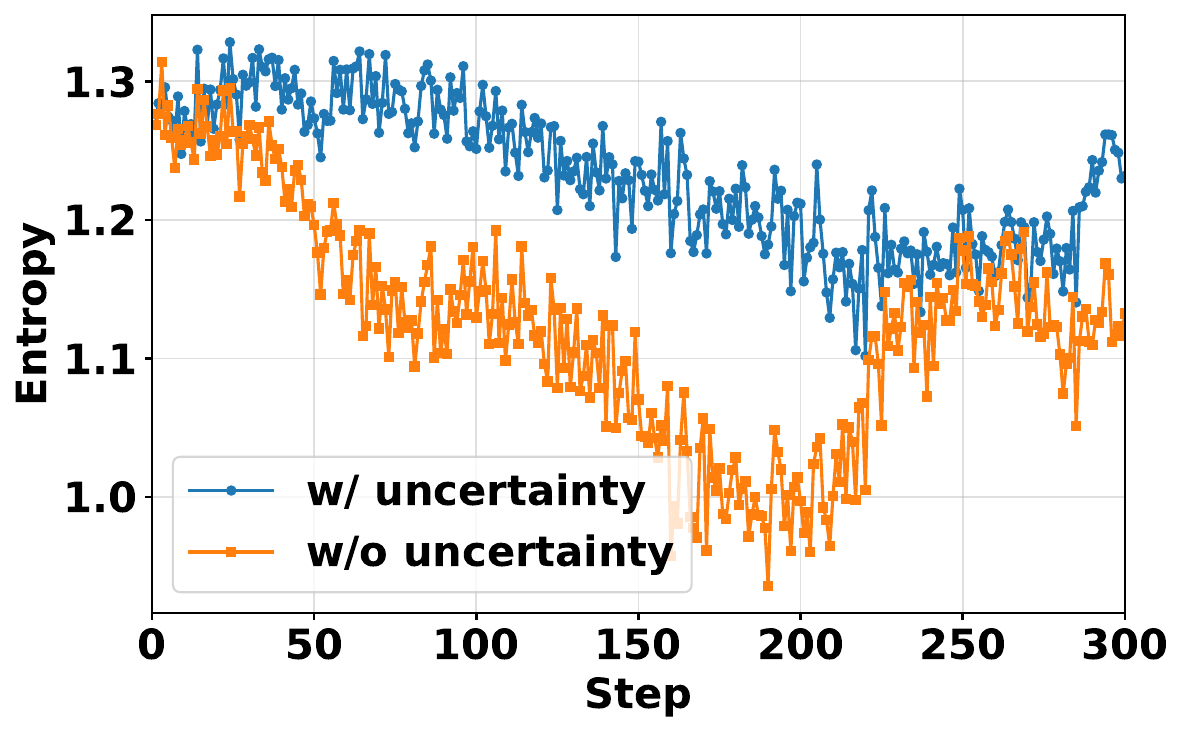}
    \vspace{-0.5cm}
    \captionof{figure}{The entropy curve with and without uncertainty-aware reweighting.}
    \label{fig:uncertainty_figure}
    \par\vspace{0.2cm}
    \includegraphics[width=\linewidth]{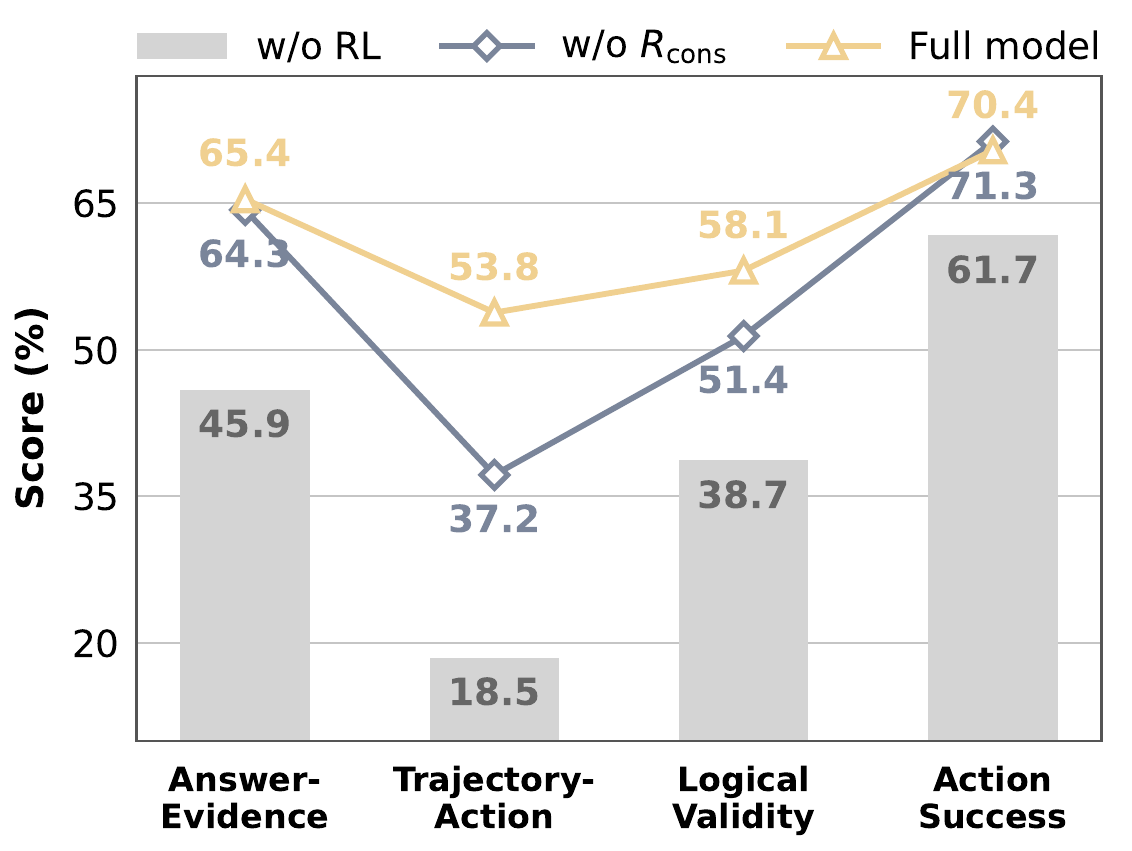}\vspace{-0.2cm}
    \captionof{figure}{Ablation of RL components on numerical grounding metrics}\label{fig:numerical_grounding_ablation}
\end{minipage}
\vspace{-0.4cm}
\end{wrapfigure}
\textbf{Uncertainty-Aware RL Stabilizes Cross-Modal Alignment.}
Table~\ref{tab:ugrpo_ablation} shows that u-GRPO achieves the best average accuracy, while removing either the action reward $R_{\mathrm{act}}$ or the forecast precision reward $R_{\mathrm{prec}}$ leads to performance degradation. The larger drop from removing $R_{\mathrm{act}}$ highlights the importance of explicitly supervising the intermediate forecast action. Figure~\ref{fig:uncertainty_figure} further shows that uncertainty-aware reweighting stabilizes RL training: without uncertainty weighting, policy entropy drops more sharply in the early stage, indicating premature convergence and reduced exploration. In contrast, uncertainty-aware reweighting maintains higher entropy by down-weighting high-volatility, low-reliability samples, which helps preserve exploration and yields stronger late-stage performance. These results suggest that the reward design improves both cross-modal credit assignment and optimization robustness under noisy market dynamics.

\textbf{Consistency Reward Improves Numerical Grounding.}
To probe how RL helps beyond task accuracy, we use LLM-based rubrics to evaluate numerical grounding across four dimensions, including answer--evidence consistency, trajectory--action alignment, logical validity, and action success rate (detailed in Appendix~\ref{app:llm-as-judge}). As shown in Figure~\ref{fig:numerical_grounding_ablation}, removing RL entirely causes the largest degradation, confirming that reinforcement learning is essential not only for answer correctness but also for aligning the model's qualitative outlook, numerical forecast, and final reasoning. More revealingly, removing only the consistency reward $R_{\mathrm{cons}}$ preserves answer--evidence consistency but disproportionately harms trajectory--action alignment, indicating that final-answer supervision alone cannot ensure the structured forecast action remains numerically compatible with the generated trajectory. The consistency reward thus provides fine-grained credit assignment at the action--trajectory interface, preventing forecast actions that are semantically plausible but numerically unsupported.



\section{Conclusion}
We introduce \name, a time-series-enhanced financial LLM framework that couples language-based market reasoning with trajectory-level numerical forecasting through executable forecast actions. Instead of requiring the LLM to directly generate dense future trajectories, \name uses forecast actions as an intermediate interface to guide a dedicated time-series decoder and to support verifiable alignment between reasoning, forecasting, and final answers. We further optimize this process with consistency-grounded RL and uncertainty-aware reweighting, improving robustness under noisy and non-stationary market dynamics. Experiments show that \name improves both financial QA and time-series forecasting, demonstrating the value of explicitly grounding LLM reasoning in modality-specific temporal prediction.

\newpage
\bibliographystyle{unsrt}
\bibliography{reference}
\newpage
\newpage
\appendix
\onecolumn
\section{Dataset Curation}\label{Appx:data_details}
\subsection{Data Collection and Preprocessing}
To construct a robust multimodal benchmark, we aggregate high-frequency market data and unstructured textual signals from distinct providers, ensuring strict temporal synchronization. 

\xhdr{Numerical Data Acquisition and Preprocessing.} 
Historical daily pricing data, covering Open, High, Low, Close, and Volume (OHLCV) for the S\&P 500 universe, is retrieved via the \texttt{yfinance} API\footnote{\url{https://ranaroussi.github.io/yfinance/}}. To mitigate the inherent non-stationarity of raw prices and facilitate efficient representation learning, we apply a log-return transformation to the raw inputs $\mathbf{X}_t$. Specifically, raw price components are normalized relative to the previous day's closing price, formally defined as $r_t^{\mathrm{N}}=\ln(\frac{\mathrm{N}_t}{C_{t-1}})$ for $\mathrm{N}\in\{O,H, L,C\}$. Similarly, the volume sequence is log-normalized as $r_t^{V}=\ln(\frac{V_t+1}{V_{t-1}+1})$ to reduce the impact of outliers.

\xhdr{Textual Data Collection and Alignment.} 
Semantic context is sourced from public market API\footnote{\url{https://massive.com/}} to complement the numerical signals. We enforce a rigorous alignment protocol to prevent look-ahead bias and ensure causal validity. For any given trading day $t$:
\begin{enumerate}[leftmargin=0.5cm,nosep]
\item  \textbf{Company Fundamentals}: We retrieve the most recently published quarterly financial statement prior to $t$ (\eg revenue, operating margins, leverage ratios).
\item  \textbf{Business \& News Context}: We aggregate news articles from a fixed lookback window using strict keyword matching based on the company's ticker and official name, retaining only information available before market close at $t-1$.
\item  \textbf{Macroeconomic Data}: Key indicators such as US Treasury yields (\eg 2Y vs. 10Y spreads) and inflation metrics are similarly aligned by selecting the latest available release prior to time $t$. 
These heterogeneous data sources are cleaned and serialized into a unified structured prompt $\mathcal{T}_t$, serving as the grounding context for the model.
\end{enumerate}

\subsection{Dataset Statistics}\label{app:data_stats}

\xhdr{Time-Series Data}
Our numerical dataset consists of daily equity time series spanning from \textbf{2015-01-01 to 2025-10-01} over the \textbf{S\&P~500} universe. For each stock, we record five standard market channels, \ie \textit{Open, High, Low, Close,} and \textit{Volume} (OHLCV), at daily resolution. After aligning trading calendars and removing missing or inactive tickers, the dataset contains approximately $2{,}700$ trading days per asset. All price-related channels are transformed into log-returns relative to the previous close to mitigate non-stationarity, while volume is log-normalized. The resulting multivariate time series serves as the numerical backbone for both forecasting and reasoning tasks.

\xhdr{Question Answering Data} On top of the aligned time-series and textual context, we construct a large-scale financial QA dataset covering five task categories: \textit{Forecast QA, Event Detection, Fundamental QA, News Analysis,} and \textit{Multi-signal Reasoning}. The dataset contains 40,000 training questions and 4,000 evaluation questions, each paired with temporally aligned numerical trajectories, macroeconomic indicators, company fundamentals, and the 10 most recent news contexts. All questions are grounded at a specific decision time $t$, with answers constrained to use only information available prior to or at $t$ to avoid look-ahead bias. Each QA instance includes a structured reasoning trace and a final answer, enabling both supervised fine-tuning and reinforcement learning for time-series-grounded reasoning.

\subsection{Cold-start Data Generation}\label{appd:cold_data}
Synthesizing high-quality supervision via model distillation is essential for bootstrapping reasoning in specialized financial domains, where generalist models often struggle to jointly integrate numerical dynamics and semantic context~\cite{liu2025fin,wang2023fingpt,shao2024deepseekmath}. Based on the unified multimodal context, we construct cold-start trajectories that follow a tool-call interface. Concretely, GPT-5 first produces a concise market analysis, then emits a structured action followed by \texttt{<forecast\_action>} that specifies forecast-relevant fields. This action is used to condition the time-series decoder, and the returned trajectory is provided to the model for final reasoning and a concise answer in \texttt{<answer>}. Our pipeline consists of three stages, each targeting a specific source of noise or inefficiency in synthetic supervision:
\begin{enumerate}[leftmargin=0.5cm,nosep]
\item \textbf{Domain-Aware Question Synthesis}.
We prompt GPT-5 to generate questions across diverse financial tasks (\eg \textit{Forecast QA}, \textit{Event Detection}, \textit{News Analysis}, and \textit{Multi-signal Reasoning}) to cover varied reasoning horizons and decision objectives.
\item \textbf{Action-grounded Reasoning Construction}.
For each query, we derive a target forecast action from the realized future trajectory and require the teacher trace to justify an action that is consistent with available historical, textual, and macroeconomic evidence. This produces supervision for the reasoning before the intermediate action that will consequently control the decoder.
\item \textbf{Tool-aware Verification and Filtering}.
We execute the action-conditioned \texttt{<forecast\_action>} to obtain a numerical trajectory, verify whether the trajectory supports the target answer, and filter out samples that are either inconsistent or trivially solvable by a smaller baseline (\eg Qwen3-4B). This yields $\mathcal{D}_{\text{SFT}}$ with explicit reasoning-action-answer alignment.
\end{enumerate}

\subsection{Examples}\label{Appx:Examples}

\begin{tcolorbox}[
  enhanced,
  breakable,
  colback=light-primary,
  colframe=dark-secondary!60!black,
  title=Examples of Questions,
  fonttitle=\bfseries,
  sharp corners,
  boxrule=0.6pt,
  left=6pt,right=6pt,top=6pt,bottom=6pt,
]

\textbf{Forecasting QA Questions:}
\begin{itemize}[nosep]
  \item What will be the maximum drawdown within the next $x$ trading days?
  \item What will be the percentage change in the closing price over the next $x$ days??
\end{itemize}

\textbf{Price-action Event Detection Questions:}
\begin{itemize}[nosep]
  \item Will the stock drop more than 5\% from today's close within the next $x$ days?
  \item Will the stock hit more than 5\% from today's close in the next $x$ days?
\end{itemize}
\textbf{Fundamental Questions:}
\begin{itemize}[nosep]
  \item Given the current 10-year vs 2-year Treasury yield spread, is the stock predicted to have higher volatility or lower volatility in the next 10 days?
  \item If the 1-month yield exceeds the 10-year yield (yield curve inversion), will the stock's next-10-day cumulative return stay positive?
\end{itemize}
\textbf{News Analysis Questions:}
\begin{itemize}[nosep]
  \item Given the 10 most recent headlines, will the stock's predicted 7-day cumulative return stay positive?
  \item If most news sentiment is negative, will the stock still recover above today's price within 10 days?
\end{itemize}

\textbf{Multi-signal Reasoning Questions:}
\begin{itemize}[nosep]
  \item Given the current market conditions, will the stock's predicted volatility rise or fall?
  \item Combining macro shifts, news flow, and price forecasts, will you recommend a scale-in, scale-out, or stay-flat strategy?
\end{itemize}

\end{tcolorbox}

\begin{tcolorbox}[
  enhanced,
  breakable,
  colback=green-light-primary,
  colframe=green-light-secondary!60!black,
  title=Examples of Multi-signal Input,
  fonttitle=\bfseries,
  sharp corners,
  boxrule=0.6pt,
  left=6pt,right=6pt,top=6pt,bottom=6pt,
]
\textcolor{green-light-secondary!60!black}{\textbf{Ticker:}} XYZ

\textcolor{green-light-secondary!60!black}{\textbf{Date:}} 2025-06-20

\textcolor{green-light-secondary!60!black}{\textbf{Company:}} Block, Inc. (XYZ)

\textcolor{green-light-secondary!60!black}{\textbf{Business Summary:}} Founded in 2009, Block provides payment services to merchants, along with related services. The company also launched Cash App, a person-to-person payment network. In 2024, Square's payment volume was almost USD 250 million.

\noindent \textcolor{green-light-secondary!60!black}{\textbf{Macro Snapshot:}}
\begin{itemize}[nosep]
    \item US 1M Treasury yield: 4.20\%
    \item US 3M Treasury yield: 4.39\%
    \item US 2Y Treasury yield: 3.90\%
    \item US 10Y Treasury yield: 4.38\%
    \item US 30Y Treasury yield: 4.89\%
\end{itemize}

\noindent \textcolor{green-light-secondary!60!black}{\textbf{Fundamentals (Q1 2025):}}
\begin{itemize}[nosep]
    \item revenue: 5.77B
    \item gross profit: 2.29B
    \item operating income: 329.30M
    \item net income: 188.72M
    \item gross margin: 40.00\%
    \item operating margin: 6.00\%
    \item net margin: 3.00\%
    \item operating cash flow margin: 2.00\%
    \item total assets: 36.40B
    \item total liabilities: 14.96B
    \item shareholders' equity: 21.44B
    \item current ratio: 2.27
    \item quick ratio: 2.26
    \item liabilities-to-equity: 0.70
    \item debt-to-assets: 0.41
    \item working capital: 11.03B
    \item asset turnover: 0.16
    \item effective tax rate: 17.00\%
    \item cash from operations: 133.34M
    \item cash from investing: 914.71M
    \item cash from financing: -1.21B
    \item net cash flow: -141.34M
\end{itemize}

\noindent \textcolor{green-light-secondary!60!black}{\textbf{News:}}
\begin{itemize}[nosep]
    \item \textbf{2025-06-08: The 3 Best Growth Stocks to Buy With \$100 Right Now} \\
    The article discusses three growth stocks that have been beaten down by the market but could produce incredible returns going forward: Marvell Technology, Block, and DraftKings. It highlights the growth prospects and potential value of these companies. \\
    \textbf{Sentiment:} positive \\
    \textbf{Sentiment Reason:} The article believes Block's Cash App and Square businesses have long-term growth potential, and the recent sell-off in the stock price presents a good opportunity for growth investors.

    \item \textbf{2025-05-31: Better Growth Stock: Block vs. SoFi Technologies} \\
    The article compares two fintech companies, SoFi Technologies and Block (formerly Square), and suggests that SoFi has a more promising future than Block. SoFi is benefiting from the growing trend of online banking, while Block's growth is slowing due to increased competition in the payment processing space and uncertainty around the adoption of blockchain technology. \\
    \textbf{Sentiment:} negative \\
    \textbf{Sentiment Reason:} The article suggests that Block's growth is slowing due to increased competition in the payment processing space and uncertainty around the adoption of blockchain technology, which the company has prioritized.

    \ldots
\end{itemize}
\end{tcolorbox}

\section{Method Details}
\subsection{Time-series Encoder}\label{appd:ts_encoder}
\xhdr{Additional Details on Multi-Signal Time-Series Encoder}
We provide a more explicit formulation of the multi-signal time-series encoder used in Sec.~\ref{sec:ts_encoder_pretrain}.
Given a multichannel input sequence $\mathbf{X}_t \in \mathbb{R}^{C \times L}$ with channel index $v \in \{1,\dots,C\}$ and temporal index $\tau \le t$, we flatten the sequence into a token stream while preserving temporal order.
Each token embedding is augmented with Rotary Positional Embeddings (RoPE), where the rotation angle is a deterministic function of the absolute timestamp $\tau$, ensuring that relative temporal offsets are encoded directly in the query--key inner product.
Formally, for attention head $h$, the rotated queries and keys are given by
$
\tilde{\mathbf{q}}_{\tau}^{(h)} = \mathrm{RoPE}(\mathbf{q}_{\tau}^{(h)}, \tau), \quad
\tilde{\mathbf{k}}_{\tau}^{(h)} = \mathrm{RoPE}(\mathbf{k}_{\tau}^{(h)}, \tau)
$.

To enforce causal consistency across all variables, we introduce a block-diagonal causal mask $M \in \{0, -\infty\}^{(CL)\times(CL)}$, where
\[
M_{(\tau,v),(\tau',v')} =
\begin{cases}
0, & \tau' \le \tau, \\
-\infty, & \text{otherwise}.
\end{cases}
\]
This mask guarantees that representations at time $\tau$ only attend to historical observations from all channels.
In addition, we inject a learnable binary bias matrix $B$ into the attention logits to distinguish intra-variable dependencies from inter-variable correlations:
\[
B_{(\tau,v),(\tau',v')} =
\begin{cases}
b_{\text{intra}}, & v = v', \\
b_{\text{inter}}, & v \neq v',
\end{cases}
\]
where $b_{\text{intra}}$ and $b_{\text{inter}}$ are trainable scalars shared across layers.
The resulting attention computation for head $h$ is
\[
\mathrm{Attn}^{(h)}(\tau,v) =
\sum_{\tau',v'} \mathrm{softmax}\!\left(
\frac{\tilde{\mathbf{q}}_{\tau,v}^{(h)} \cdot \tilde{\mathbf{k}}_{\tau',v'}^{(h)}}{\sqrt{d_h}}
+ M_{(\tau,v),(\tau',v')}
+ B_{(\tau,v),(\tau',v')}
\right)\mathbf{v}_{\tau',v'}^{(h)}.
\]

The encoder outputs a sequence of latent time-series tokens $\mathbf{H}_{\mathrm{ts}} \in \mathbb{R}^{K \times d_m}$, which are subsequently projected to parametric distribution heads.
For each forecasting horizon $s \in \{1,\dots,S\}$ and target variable $i \in \{O,C,V\}$, we obtain the mean and variance parameters via linear projections:
\[
\mu_{t+s}^{(i)} = \mathbf{w}_{\mu}^{(i)\top}\mathbf{h}_{t}^{\mathrm{ts}}, \quad
\sigma_{t+s}^{(i)} = \mathrm{softplus}\!\left(\mathbf{w}_{\sigma}^{(i)\top}\mathbf{h}_{t}^{\mathrm{ts}}\right),
\]
where $\mathbf{h}_{t}^{\mathrm{ts}}$ denotes the aggregated latent state corresponding to time $t$.
The softplus activation ensures positivity of $\sigma$ and stabilizes training.
For high and low prices, separate heads predict quantile estimates $q_{t+s}^{(H,\tau_H)}$ and $q_{t+s}^{(L,\tau_L)}$, enabling asymmetric modeling of upside and downside risk.
Together, this design allows the encoder to produce both point forecasts and calibrated uncertainty estimates, which are later consumed by the forecasting decoder and the uncertainty-aware reinforcement learning stage.

\subsection{Structured Forecast Actions}\label{appx:forecast_actions}
\begin{tcolorbox}[
  enhanced,
  colback=gray!5,
  colframe=green!20!black,
  sharp corners,
  boxrule=0.3pt,
  left=5pt,right=5pt,top=5pt,bottom=5pt,
]
\begin{verbatim}
{
  "future_window": "t+1_t+10",
  "start_value": <float>,
  "end_value": <float>,
  "max_value": <float>,
  "min_value": <float>,
  "mean_close": <float>,

  "direction": "<up | down | flat>",
  "end_change_pct": <float>,
  "range_pct": <float>,
  "volatility_level": "<low | medium | high>",
  "range_width_bin": "<narrow | moderate | wide>",
  "max_drawdown_bin": "<low | medium | high>",

  "turning_point_count": <int>,
  "peak_timing_bin": "<early | middle | late>",
  "trough_timing_bin": "<early | middle | late>",
  "monotonicity": "<increasing | decreasing | mixed>",
  "trendline_fit": "<weak | moderate | strong>",
  "tail_risk_level": "<low | medium | high>"
}
\end{verbatim}
\end{tcolorbox}
Each action is implemented as a JSON-style dictionary containing three groups of fields: numerical summary statistics of the future trajectory, categorical or ordinal market states, and temporal shape descriptors. During supervised data construction, the target action is deterministically derived from the realized future trajectory, providing field-level supervision for the LLM's intermediate forecast decision. During inference, the LLM emits the same structured action before invoking \texttt{<forecast\_action>}; the action fields are embedded and passed through an MLP action encoder to produce the conditioning vector for the time-series decoder. This design makes the LLM's market outlook explicit, executable, and field-wise verifiable.

\subsection{Reinforcement Learning Preliminary}\label{appx:grpo}
Our RL training is based on GRPO algorithm~\cite{shao2024deepseekmath}, which is a variant of policy-gradient optimization designed for large language models, where per-query baselines are estimated from a group of sampled rollouts rather than a learned value function.

\xhdr{Rollout Generation}
Given an input query $q$, the current policy $\pi_\theta$ generates a group of $G$ independent rollouts $\{ o_i \}_{i=1}^G$, where each rollout $o_i = (a_{i,1}, \ldots, a_{i,T_i})$ corresponds to a complete response trajectory (\eg reasoning steps and final answer).
Each rollout is evaluated by a scalar reward function $R_i = R(o_i)$, which may incorporate correctness, formatting, or task-specific criteria.

\xhdr{Group-Relative Advantage}
Instead of estimating a value baseline, GRPO computes a group-relative advantage by normalizing rewards within each rollout group: $\hat{A}_i =\frac{R_i-\text{Mean}(\mathbf{R})}{\text{Std.}(\mathbf{R})} $.
This formulation encourages rollouts that perform better than the group average while suppressing uniformly poor generations, providing a low-variance baseline without additional critic training.

\xhdr{Optimization Objective}
GRPO adopts a PPO-style clipped objective applied at the token level. For each rollout $o_i$, the objective is defined as
\begin{equation}
\mathcal{L}_{\text{GRPO}}(\theta)
=
\mathbb{E}_{q, \{o_i\}}
\left[
\frac{1}{G}
\sum_{i=1}^G
\frac{1}{|o_i|}
\sum_{t=1}^{|o_i|}
\min\!\left(
r_{i,t}(\theta)\hat{A}_i,\;
\text{clip}(r_{i,t}(\theta), 1-\epsilon, 1+\epsilon)\hat{A}_i
\right)
-
\beta\,\mathrm{KL}\!\left(\pi_\theta \,\|\, \pi_{\text{ref}}\right)
\right],
\end{equation}
where $
r_{i,t}(\theta)
=
\frac{\pi_\theta(a_{i,t} \mid q, a_{i,<t})}
{\pi_{\text{ref}}(a_{i,t} \mid q, a_{i,<t})}
$, is the likelihood ratio between the current policy and a frozen reference policy $\pi_{\text{ref}}$.
The KL penalty regularizes policy updates and stabilizes training, while the clipping parameter $\epsilon$ prevents excessively large gradient steps. In this work, GRPO provides the foundation upon which we introduce uncertainty-aware reweighting to better handle stochastic and noisy financial environments (Sec.~\ref{sec:rl}).

\subsection{Uncertainty Reweighting}\label{appx:rl_details}
\xhdr{Uncertainty Weighting Formulation} To modulate the learning signal based on market volatility, we calculate the sample weight $w(\mathcal{U}_q)$ using an exponential decay function anchored to a stability threshold $u_0$. The formulation is defined as:
\begin{equation}\label{eq:uncertainty}
w(\mathcal{U}_q) = \begin{cases}1.0 & \text{if } \mathcal{U}_q \leq u_0 \\exp\left(-\alpha \cdot \frac{\mathcal{U}_q - u_0}{u_{\text{hi}} - u_0}\right) & \text{if } \mathcal{U}_q > u_0
\end{cases}
\end{equation}
where $\alpha$ is a hyperparameter controlling the decay rate, and $u{\text{hi}}$ represents the upper bound used for normalization.

\section{Experiments}
\subsection{Experimental Settings}\label{appx:baselines}

To rigorously evaluate the performance of our proposed method, we benchmark against state-of-the-art time series forecasting models, including training-from-scratch ones and foundation models.

\paragraph{Evaluation Metrics for Time-Series Forecasting.}
For the $n$-th instance, let the historical close at the last observed day be $c_n^{(0)}$ and the future close trajectory over horizon $H=10$ be $\{c_{n,h}\}_{h=1}^{H}$, with predictions $\{\hat c_{n,h}\}_{h=1}^{H}$. We report (i) \textbf{Price} errors: mean absolute percentage error (MAPE) and root mean squared error (RMSE).
\begin{equation}
\mathrm{MAPE}_{\text{price}}=\frac{1}{NH}\sum_{n=1}^{N}\sum_{h=1}^{H}\left|\frac{\hat c_{n,h}-c_{n,h}}{c_{n,h}}\right|, \quad \mathrm{RMSE}_{\text{price}}=\sqrt{\frac{1}{NH}\sum_{n=1}^{N}\sum_{h=1}^{H}\big(z(\hat c_{n,h})-z(c_{n,h})\big)^2},
\end{equation}
where $z(\cdot)$ is instance-wise normalization (applied per sample across the horizon to reduce scale effects).

(ii) \textbf{Return} metrics based on simple returns relative to the historical close (with $\epsilon>0$ for numerical stability):
\begin{equation}
r_{n,h}=\frac{c_{n,h}}{\max(c_n^{(0)},\epsilon)}-1,\qquad 
\hat r_{n,h}=\frac{\hat c_{n,h}}{\max(c_n^{(0)},\epsilon)}-1,
\end{equation}
and daily-return RMSE across all horizons,
\begin{equation}
\mathrm{RMSE}_{\text{return}}=\sqrt{\frac{1}{NH}\sum_{n=1}^{N}\sum_{h=1}^{H}\big(\hat r_{n,h}-r_{n,h}\big)^2}.
\end{equation}
Directional accuracy at horizon $k\in\{3,5,10\}$ uses cumulative return $\hat R_{n}^{(k)}=\hat r_{n,k}$ and $R_{n}^{(k)}=r_{n,k}$:
\begin{equation}
\mathrm{DA}_{k}=\frac{1}{N}\sum_{n=1}^{N}\mathbb{I}\Big[\mathrm{sign}(\hat R_{n}^{(k)})=\mathrm{sign}(R_{n}^{(k)})\Big].
\end{equation}

(iii) \textbf{Volatility} is evaluated by realized volatility over the predicted/true close paths as the sum of squared log-returns:
\begin{equation}
\begin{aligned}
    \mathrm{RV}_n&=\sum_{h=1}^{H-1}\Big(\log\max(c_{n,h+1},\epsilon)-\log\max(c_{n,h},\epsilon)\Big)^2,\\
\widehat{\mathrm{RV}}_n&=\sum_{h=1}^{H-1}\Big(\log\max(\hat c_{n,h+1},\epsilon)-\log\max(\hat c_{n,h},\epsilon)\Big)^2.
\end{aligned}
\end{equation}
We report $\mathrm{MAE}_{\text{RV}}=\frac{1}{N}\sum_{n=1}^{N}\big|\widehat{\mathrm{RV}}_n-\mathrm{RV}_n\big|$ and the coefficient of determination.

(iv) \textbf{Volume} metrics follow the same horizon-based aggregation on the future volume series $\{v_{n,h}\}$ and predictions $\{\hat v_{n,h}\}$, using instance-wise normalization $z(\cdot)$:
\begin{equation}
\begin{aligned}
\mathrm{MAPE}_{\text{vol}}&=\frac{1}{NH}\sum_{n=1}^{N}\sum_{h=1}^{H}\left|\frac{z(\hat v_{n,h})-z(v_{n,h})}{\max(z(v_{n,h}),\epsilon)}\right|,\\
\mathrm{RMSE}_{\text{vol}}&=\sqrt{\frac{1}{NH}\sum_{n=1}^{N}\sum_{h=1}^{H}\big(z(\hat v_{n,h})-z(v_{n,h})\big)^2}.
\end{aligned}
\end{equation}

\subsection{Hyper-parameter Settings}\label{appd:hyper_setting}
For time-series encoder pretraining, we use a 90-day input window to predict the next 10 trading days. The encoder adopts a patch length of 5 with dropout set to 0.1. We sweep model capacities by varying the hidden dimension $d_{\text{model}}\in\{512,768,1024\}$, paired with $\{8,12,16\}$ attention heads and $\{10,16\}$ Transformer layers. All models are trained for 50 epochs using AdamW with learning rate $1\times10^{-4}$ and weight decay $0.01$, a cosine learning-rate schedule with 1000 warmup steps. Unless otherwise stated, pretraining uses market data spanning 2015-01-01 to 2025-01-01 and test from 2025-01-01 to 2025-10-01. Table~\ref{tab:encoder_config} summarizes the architectural configurations of the time-series encoders at different scales.

\begin{table}[ht]
\centering
\caption{Architecture configurations of the time-series encoder at different scales.}
\label{tab:encoder_config}
\resizebox{0.6\textwidth}{!}{%
\begin{tabular}{lcccc}
\toprule
\textbf{Encoder Size} & \textbf{Hidden Dim} & \textbf{\# Layers} & \textbf{\# Attention Heads} &\textbf{\# Parameters} \\
\midrule
Small  & 512  & 8  & 8 & 27M\\
Base   & 768  & 12 & 12 &65M\\
Large  & 1024 & 16 & 16 & 119M\\
\bottomrule
\end{tabular}}
\end{table}

For supervised fine-tuning (SFT), we fine-tune \texttt{Qwen3-4B-Instruct-2507} and \texttt{Qwen3-8B} using LoRA with rank $r=32$, scaling factor $\alpha=32$, applied to projection and feed-forward modules, and initialize the model with pre-trained time-series encoder checkpoints. Supervised fine-tuning is conducted with the AdamW optimizer with learning rate $2\times10^{-5}$ and weight decay 0.01, cosine warmup of 100 steps for 5 epochs.

For reinforcement learning, we employ VERL with uncertainty-aware reweighting of the learning signal. We select the reward weights through validation search and use $\alpha=1.0$, $\beta=0.5$, and $\gamma=1.0$ as the default setting, which empirically yields the best trade-off among action accuracy, forecast precision, and action--trajectory consistency. The uncertainty parameters are set to $u_0=0.25$, $u_{\text{hi}}=0.50$, $w_{\text{hi}}=0.50$, and $u_{\text{cap}}=1.0$ in Eq.~\ref{eq:uncertainty}. Group size $G$ is tuned over $\{4, 8, 12\}$ and the default configuration is set to $G=8$.

\subsection{Effect of Uncertainty and Rollout Size}\label{app:uncertainty}
\begin{figure}[t]
    \centering
    \begin{minipage}[t]{0.49\columnwidth}
        \centering
        \includegraphics[width=\linewidth]{figures/entropy_vs_steps.pdf}
    \end{minipage}
    \begin{minipage}[t]{0.40\columnwidth}
        \centering
        \includegraphics[width=\linewidth]{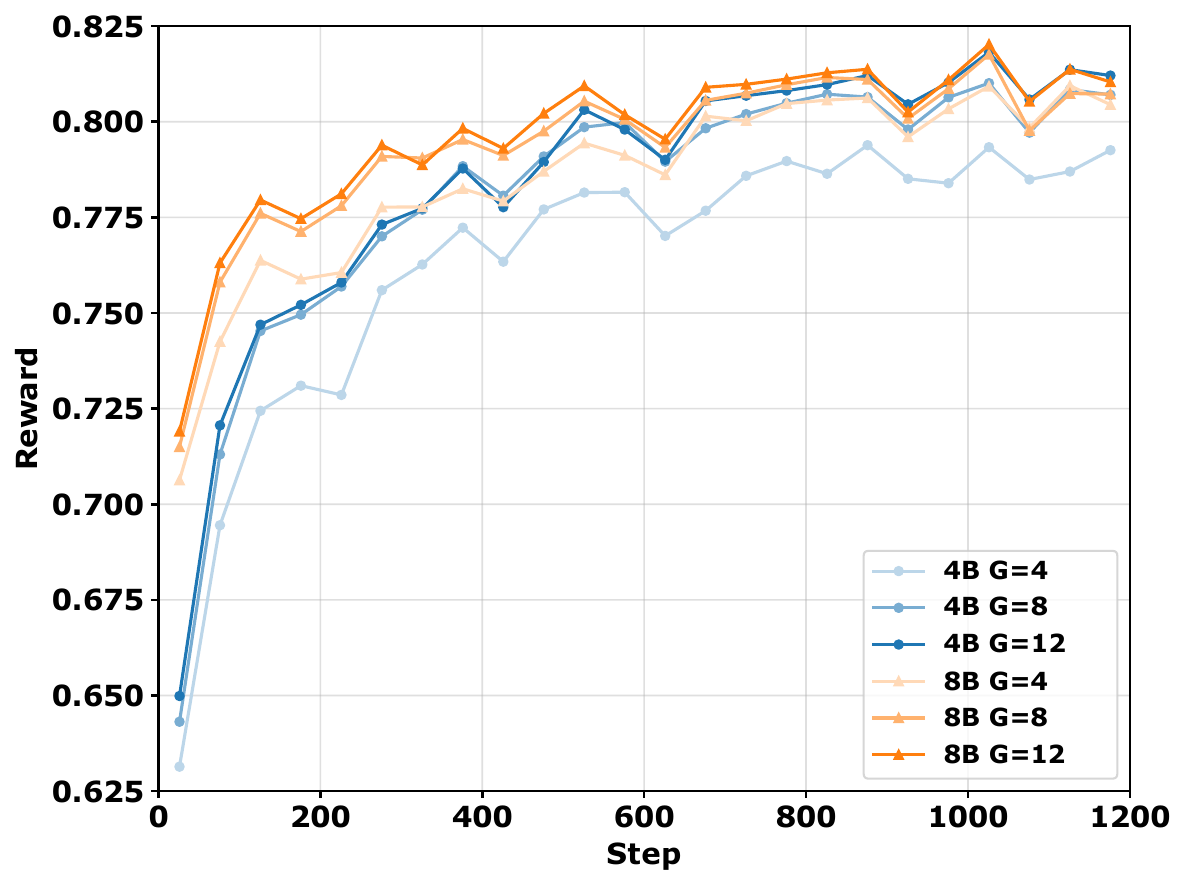}\vspace{-0.1cm}
    \end{minipage}
    \caption{\textbf{Left:} Policy entropy over training steps, comparing training with and without uncertainty-aware reweighting. \textbf{Right:} Reward convergence under different group sizes.}\vspace{-0.2cm}
    \label{fig:uncertainty}
\end{figure}

Figure~\ref{fig:uncertainty} (Left) illustrates the role of uncertainty-aware reweighting in stabilizing training dynamics and improving downstream performance. As shown in the left panel, training without uncertainty leads to a faster and more pronounced decrease in policy entropy in the early stage of RL training, indicating premature loss of exploration and increasingly deterministic behavior. This entropy collapse is also accompanied by higher instability across training steps. Instead, uncertainty-aware reweighting maintains consistently higher entropy, encouraging sustained exploration by down-weighting high-volatility and low-reliability samples. This difference in learning dynamics translates into better late-stage performance: while the model trained without uncertainty saturates early, the uncertainty-aware reweighting continues to improve and achieves higher accuracy ultimately. These results suggest that uncertainty-aware reweighting mitigates premature convergence, preserves exploration under noisy market regimes, and leads to more robust policy optimization.

Figure~\ref{fig:uncertainty} (Right) studies the effect of the group size $G$ on training reward convergence for both \name-4B and \name-8B. Across both model scales, we observe that increasing $G$ generally improves reward stability and final convergence, as larger groups provide a more reliable estimate of relative advantages. The impact is more pronounced for the smaller 4B model, indicating higher variance in policy updates and suboptimal optimization. In contrast, the 8B model is more robust to smaller group sizes, though it still benefits from moderate increases in $G$. To balance optimization quality and training efficiency, we therefore select $G=8$ as the default rollout number.

\subsection{Scalability Analysis} \label{appd:scale_ablation}
\begin{wrapfigure}{r}{0.55\textwidth}\vspace{-0.5cm}
  \centering
    \includegraphics[width=\linewidth]{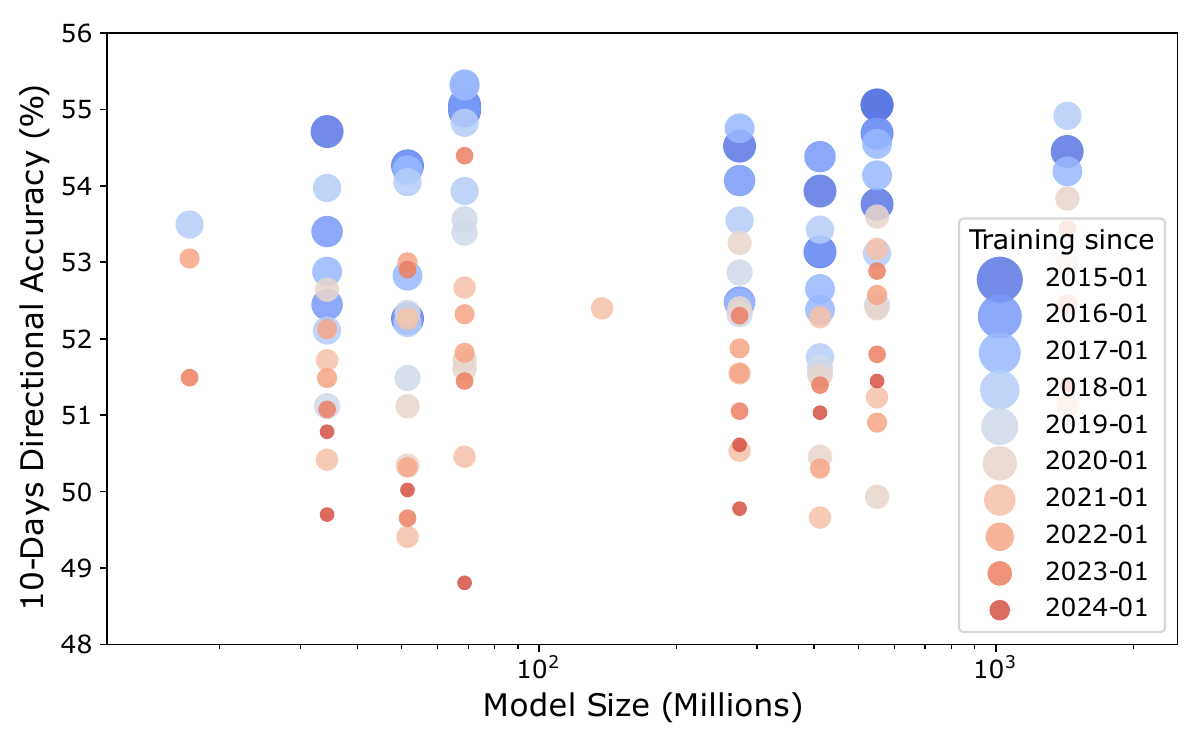}\vspace{-0.3cm}
    \caption{Scalability analysis of pretraining performance w.r.t model size (log scale) and training data.}
    \label{fig:scalability}\vspace{-0.5cm}
\end{wrapfigure} 
We investigate the scalability of the proposed time-series encoder along two critical dimensions: pre-training data size and model capacity. As illustrated in Figure \ref{fig:scalability}, we systematically vary the training data duration by shifting the start date from 2015 to 2024, while keeping the test set fixed. Simultaneously, we scale the model size from approximately 20M to 2B parameters by adjusting the depth and width of the Transformer layers. First, we observe an obvious performance stratification regarding data scalability: models trained on longer historical horizons (represented by cooler colors, \eg 2015-2016) consistently outperform those trained on limited recent data (warmer colors), validating that extensive historical context helps capture diverse market regimes. Second, regarding model scalability, we observe a growth in performance that saturates at a specific capacity threshold at approximately $10^2$M parameters, indicating that the size is sufficiently robust for uni-modal time-series pretraining, balancing high predictive accuracy with computational efficiency before the integration of multimodal reasoning layers. 

\begin{table}[ht]
\vspace{-6pt}
\centering
\caption{QA performance of \name-4B with different sizes of time-series encoder.}
\label{tab:QA_scale}
\resizebox{\textwidth}{!}{%
\begin{tabular}{llcccccc}
\toprule
Setting & Stage & Pure Forecast & Event Detection & Macro Fundamental & News Sentiment & Multi-Signal Reasoning & Overall \\
\midrule
\multirow{2}{*}{4B - Small} 
& SFT & 37.04 & 63.26 & 50.78 & 48.99 & 58.05 & 52.00 \\
& SFT + u-GRPO & 48.11 & 59.18 & 48.80 & 60.09 & 71.07 & 57.50 \\
\midrule
\multirow{2}{*}{4B - Base} 
& SFT & 45.62 & 65.75 & 48.61 & 56.64 & 65.80 & 56.50 \\
& SFT + u-GRPO & 51.00 & 68.52 & 59.70 & 51.83 & 63.02 & 59.00 \\
\midrule
\multirow{2}{*}{4B - Large} 
& SFT & 41.62 & 58.12 & 60.00 & 57.37 & 60.36 & 55.60 \\
& SFT + u-GRPO & 50.28 & 60.22 & 61.38 & 55.49 & 63.72 & 58.21 \\
\bottomrule
\end{tabular}}
\vspace{-8pt}
\end{table}
Table~\ref{tab:QA_scale} examines how encoder capacity interacts with the underlying LLM size in downstream financial QA tasks. We do not observe a clear monotonic trend with respect to encoder scale. While moderate encoder capacity performs competitively, further scaling does not consistently translate into additional gains. Overall, these results highlight the importance of considering encoder--LLM capacity matching rather than indiscriminately increasing model size.

\subsection{Investment Simulation}\label{sec:simulation}
\xhdr{Backtest Setting} To assess whether improved forecasting performance translates into economically meaningful decisions, we further evaluate all time-series models in a unified investment simulation. We design a standardized backtesting strategy that converts each model’s 10-day close-price forecast into discrete buy or sell signals. At each trading step, the predicted cumulative return is normalized by recent realized volatility to form a z-score signal. When the signal exceeds a fixed threshold, the strategy enters a long or short position with a fixed capital allocation and holds it for a fixed horizon. This protocol enforces identical trading rules across all models, ensuring that return differences stem from forecasting quality rather than strategy engineering.

\xhdr{Results} Table~\ref{tab:backtest} reports portfolio-level metrics over U.S. equities spanning multiple market sectors, including technology (\eg NVDA), finance (\eg JPM), consumer (\eg WMT), and broad-market indices  (\eg SPY). Overall, \name achieves the strongest risk-adjusted performance, with the highest Sharpe ratio and annualized return among all baselines. While exhibiting a higher market beta, \name maintains a competitive drawdown, indicating that its increased exposure to the market is driven by more confident and selective trend exploitation rather than excessive risk-taking. In contrast, several classical time-series baselines yield unstable or even negative Sharpe ratios.
\begin{wraptable}{r}{0.7\textwidth}
\centering
\caption{Backtesting (from 2025-01-01 to 2025-10-01) performance of forecasting models  (\textbf{SP}: Sharpe Ratio, \textbf{AR}: Annualized Return, \textbf{MDD}: Maximum Drawdown, \textbf{TT}: Total Traders, \textbf{WR}: Win Rate)}
\label{tab:backtest}
\resizebox{0.7\columnwidth}{!}{%
\begin{tabular}{lccccccc}
\toprule
\textbf{Model} & \textbf{SP} & \textbf{AR} & \textbf{MDD} & \textbf{Beta} & \textbf{TT} & \textbf{WR} (\%) \\
\midrule
ARIMA & 0.455 & -2.78 & 16.15 & 0.080 & 12.33 & 51.03 \\
Prophet & -0.813 & 0.56 & 11.83 & 0.239 & 15.56 & 41.39 \\
NSTransformer & 0.414 & 8.67 & 8.26 & 0.204 & 13.22 & 48.38 \\
TimesNet & 0.027 & 11.22 & 7.60 & 0.280 & 13.56 & 47.44 \\
iTransformer & 0.873 & 10.00 & 8.21 & 0.308 & 13.67 & \textbf{61.21} \\
PatchTST & 0.976 & 15.67 & \textbf{7.19} & 0.221 & 13.00 & 51.25 \\
TimeMixer & 0.705 & 17.78 & 7.85 & 0.305 & 13.11 & 50.67 \\
DLinear & -0.266 & 20.78 & 10.20 & 0.134 & 15.11 & 37.67 \\
Mamba & -0.008 & 25.22 & 8.57 & 0.400 & 13.33 & 47.42 \\
\rowcolor{lightblue!60}\name & \textbf{1.059} & \textbf{31.89} & 12.11 & \textbf{0.521} & 14.00 & 59.48 \\
\bottomrule
\end{tabular}}\vspace{-0.4cm}
\end{wraptable}
We also observe distinct behavioral differences in decision patterns across models. For instance, DLinear~\cite{nie2022time} frequently exhibits a contrarian pattern, tending to go short during rising markets and go long during drawdowns, which leads to systematic trend misalignment. In comparison, \name produces more market-aware decisions, selectively increasing exposure when predictive signals are strong and remaining conservative under ambiguous conditions. This behavior is reflected in its higher win rate and superior Sharpe ratio, suggesting that the model predicts more decision-relevant signals beyond short-term price fluctuations. 

\begin{figure}[h]
    \centering
    \begin{minipage}{\linewidth}
        \centering
        \includegraphics[width=\linewidth]{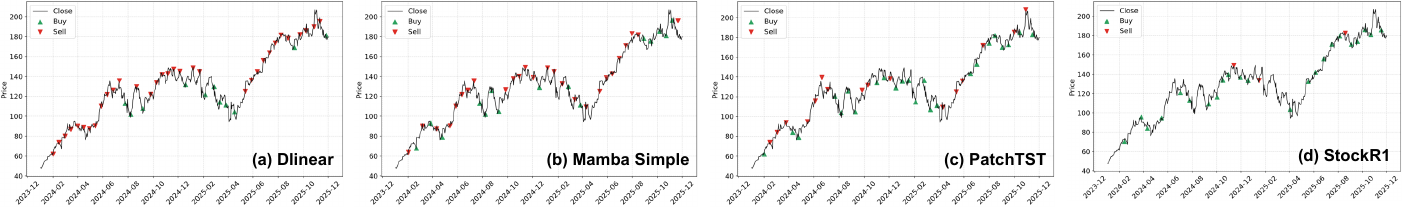}
        \caption*{(1) Directional Trading Signal Prediction for \$NVDA}
    \end{minipage}
    \begin{minipage}{\linewidth}
        \centering
        \includegraphics[width=\linewidth]{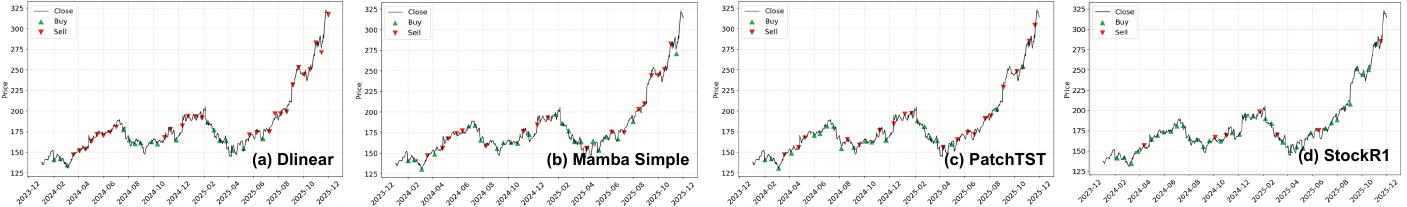}
        \caption*{(b) Directional Trading Signal Prediction for \$GOOGL}
    \end{minipage}
    \begin{minipage}{\linewidth}
        \centering
        \includegraphics[width=\linewidth]{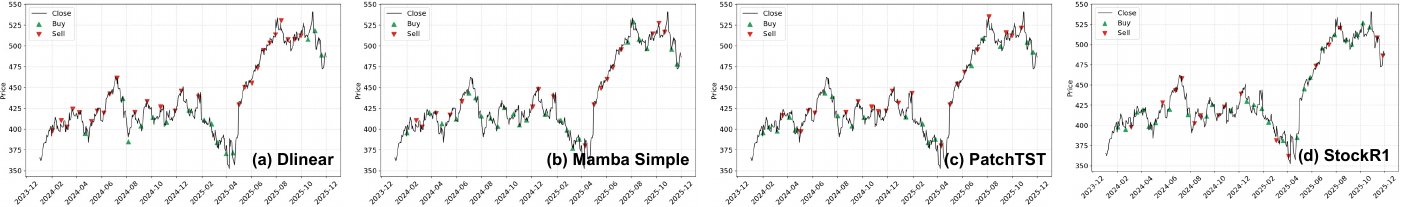}
        \caption*{(c) Directional Trading Signal Prediction for \$MSFT}
    \end{minipage}
    \caption{Directional Trading Signal Generated by Different Models}
    \label{fig:simulation}
\end{figure}
Figure~\ref{fig:simulation} provides a qualitative comparison of trading signals generated by different forecasting models under the unified backtesting protocol described in Sec.~\ref{sec:simulation}. Although all models are evaluated using the same signal extraction and execution rules, their resulting buy/sell behaviors exhibit markedly different temporal patterns, reflecting distinct implicit decision logics. Classical linear and decomposition-based models (\eg DLinear) exhibit highly reactive and oscillatory signals, often trading against prevailing trends, indicating short-horizon extrapolation and poor trend awareness. More expressive neural forecasters (\eg PatchTST) generate smoother, trend-following signals but tend to maintain persistent exposure across regime shifts due to point-forecast objectives that lack explicit uncertainty modeling. In contrast, \name produces more selective and market-aware signals, increasing exposure primarily when predictive signals are strong and remaining conservative under ambiguous or volatile conditions. As a result, \name achieves superior risk-adjusted performance (Table~\ref{tab:backtest}), indicating that its advantage stems from improved decision relevance rather than trading strategy engineering. We acknowledge that accurately predicting short-term price movements remains inherently challenging in financial markets; nevertheless, our goal is not perfect directional accuracy, but to provide more informative and risk-aware buy/sell signals that better support practical decision-making.

\subsection{Case Studies} \label{appd:case_study}
As shown in the case study, the model exhibits strong high-level consistency across its reasoning trace, structured forecast action, and decoded time series. The reasoning identifies sustained selling pressure and repeated failure near the $87$ - $88$ resistance range, which is reflected in the forecast action as \texttt{direction = down}. The time-series decoder further supports this outlook by predicting a decline in closing price by Day~7. Although minor numerical discrepancies may still exist between the qualitative reasoning and decoded trajectory, their overall directional and semantic alignment provides sufficient grounding for the model to reach the correct answer.
\begin{tcolorbox}[
  enhanced,
  breakable,
  colback=blue!3,
  colframe=blue!40!black,
  title=A Case Study of Model Internal Consistency,
  fonttitle=\bfseries,
  sharp corners,
  boxrule=0.6pt,
  left=6pt,right=6pt,top=6pt,bottom=6pt,
]
\textbf{Ticker}: SBUX, \textbf{Date}: 2025-05-15

\textbf{Question}:  Will the predicted 7-day cumulative return stay positive?

\medskip
\textbf{Response}: 

\smallskip
To answer this, I need to first look at and predict how things might evolve in the future.

SBUX has been under sustained selling pressure: from the mid-\$90s in late March/early April the stock has declined steadily to the current level around \$86, and momentum indicators remain weak. The recent bounce from \$81 appears to be a relief rally within a broader downtrend rather than a true reversal, as the stock has failed to reclaim the \$87--88 resistance zone on declining volume.

\smallskip
Fundamentally, the turnaround remains uncertain. Q2 2025 results show solid revenue growth but thin margins (7\% operating, 4\% net) and a highly leveraged balance sheet (liabilities-to-equity $\sim$-5, current ratio 0.64). The May 12 article highlights ``consumer pressure'' and ``turnaround challenges,'' and the May 8 piece frames Starbucks as a buy only if investors are willing to wait for a long-term recovery, implying near-term downside risk.

\smallskip
Given this backdrop, the stock is likely to break below current support in the first few sessions, testing the low-\$80s to high-\$70s area as sellers reassert control. A trough near \$77--78 is plausible around mid-window, followed by a partial recovery toward the low-\$80s as bargain hunters step in. However, the recovery is unlikely to reclaim the \$86 starting level within the forecast window, leaving the cumulative return negative. {\color{red}\itshape\small $\triangleright$ reasoning traces generated by the LLM}

\medskip

\begin{tcolorbox}[
  enhanced,
  colback=gray!5,
  colframe=green!20!black,
  sharp corners,
  boxrule=0.3pt,
  left=5pt,right=5pt,top=5pt,bottom=5pt,
]
\textbf{\texttt{<forecast\_action>}} {\color{red}\itshape\small $\triangleright$ structured forecast generated by the LLM}

\smallskip
\{\ttfamily\small
future\_window:\ \ t+1\_t+10\\
start\_value:\ \ \ \ 86.0\qquad end\_value:\ \ \ \ \ 83.6\\
max\_value:\ \ \ \ \ \ 84.8\qquad min\_value:\ \ \ \ \ 77.6\\
direction:\ \ \ \ \ \ down\qquad\ \ end\_change\_pct: -2.79\\
volatility:\ \ \ \ \ medium\qquad peak\_timing:\ \ \ t+8\_t+8\\
trough\_timing:\ \ t+5\_t+6\qquad monotonicity:\ \ non\_monotonic\\
mean\_close:\ \ \ \ \ 81.3\qquad\ \ tail\_risk:\ \ \ \ \ medium\\
...
\}
\end{tcolorbox}
The time-series decoder forecasts the future 10 days' prices and volume:
\begin{tcolorbox}[
  enhanced,
  colback=gray!5,
  colframe=green!20!black,
  sharp corners,
  boxrule=0.3pt,
  left=5pt,right=5pt,top=5pt,bottom=5pt,
]
\textbf{\texttt{<forecast\_ts>}} {\color{red}\itshape\small $\triangleright$ generated by a time-series decoder in a tool-call way}

\smallskip
{\ttfamily\small
Close price: [82.36, 79.73, 82.24, 80.74, 78.20, 77.57, 80.97, 84.80, 82.40, 83.63]\\
High price:\ [89.23, 84.13, 85.98, ...]\\
Low price:\ \ [81.68, 72.52, 77.86, ...]\\
Volume:\ \ \ \ [12916896, ...] 
}
\end{tcolorbox}
\medskip

To determine if the predicted 7-day cumulative return will stay positive, I need to calculate the cumulative return over the next 7 days using the forecasted close prices. The formula for cumulative return is:

((P\_\{t+7\} - P\_t) / P\_t) * 100, where P\_t is today's close and P\_\{t+7\} is the forecasted close 7 days from now.

\smallskip
Today's close: 85.94\\
Forecasted close prices for the next 7 days: [82.37, 79.73, 82.25, 80.74, 78.20, 77.57, 80.98]

\smallskip
Step 1: Calculate the cumulative return over 7 days:

{\small
Day 1:\ \ (82.37 - 85.94) / 85.94 = -0.04 (4.16\% drop) \\
Day 2:\ \ (79.73 - 82.37) / 82.37 = -0.03 (3.18\% drop) \\
Day 3:\ \ (82.25 - 79.73) / 79.73 = 0.03 (3.15\% gain) \\
Day 4:\ \ (80.74 - 82.25) / 82.25 = -0.02 (1.84\% drop) \\
Day 5:\ \ (78.20 - 80.74) / 80.74 = -0.03 (3.15\% drop) \\
Day 6:\ \ (77.57 - 78.20) / 78.20 = -0.01 (0.79\% drop) \\
Day 7:\ \ (80.98 - 77.57) / 77.57 = 0.04 (4.41\% gain) \\
}
\smallskip
Step 2: Calculate the cumulative return after 7 days:\\
((80.97 - 85.94) / 85.94) * 100 = \textbf{-5.78\%}

\smallskip
Step 3: Check if the cumulative return is positive:\\
-5.78\% is negative.

\medskip
Therefore, the predicted 7-day cumulative return will not stay positive.

\smallskip
\texttt{<answer>}\textbf{No}\texttt{</answer>}

\end{tcolorbox}

\subsection{Numerical Grounding Evaluation Rubric}\label{app:llm-as-judge}
We evaluate the numerical grounding quality of model outputs using an LLM-based judge. The judge receives each sample's question, generated reasoning, forecast hint (a structured JSON summary including start/end/max/min values, direction, and percentage change), the full forecast time series, and the model's predicted answer. Crucially, the judge is instructed to assess internal numerical faithfulness. The rubric defines four metrics, each scored continuously in $[0,1]$:

\begin{itemize}[leftmargin=1.2em, itemsep=2pt]
    \item \textbf{Answer--Evidence Consistency (M1).} Whether the final answer is consistent with the numerical evidence in the forecast time series and hint. 

    \item \textbf{Trajectory--Action Alignment (M2).} A field-level match ratio between the forecast action and the generated time series. The judge checks whether summary statistics (\eg start value, end value, max, min, direction, percentage change) faithfully reflect the underlying price arrays, with a small rounding tolerance ($\leq 0.02$).

    \item \textbf{Logical Validity (M3).} Whether the final answer, forecast action, generated time series, and reasoning are mutually consistent.

    \item \textbf{Action Success Rate (M4).} Whether the model successfully generates a complete, executable forecast action with all required fields, assessing structural completeness rather than correctness.
\end{itemize}

\noindent The full evaluation prompt, including detailed scoring guidelines and output format specifications, is available in our codebase.
\section{Discussion}

\xhdr{LLM Usage}
We use LLMs as part of the data construction, evaluation, and writing-assistance pipeline. Specifically, GPT-5 is used to synthesize cold-start reasoning trajectories under our tool-call format. We also use LLM-based rubrics to assess numerical grounding and reasoning consistency in model outputs. In addition, LLMs were used to assist with language polishing, clarity refinement, and organization of the manuscript.

\xhdr{Limitation}
This work focuses on daily-resolution equity data from S\&P 500 stocks, where aligned OHLCV signals, fundamentals, news, and macroeconomic context are relatively accessible. Extending the framework to higher-frequency trading data, broader asset classes, and longer-horizon macro-financial reasoning is a promising direction for future work. Our current forecast action design captures key trajectory-level attributes, but richer action spaces could further model uncertainty, regime shifts, and multi-asset dependencies. Future work may also explore tighter integration with portfolio-level objectives and adaptive action schemas for different financial tasks.

\xhdr{Social Impact}
Financial forecasting systems can support market analysis, risk assessment, and decision-oriented reasoning, but they should not be interpreted as guaranteed investment advice. Our framework emphasizes verifiable intermediate actions and numerical grounding, which may improve transparency compared with purely text-based financial reasoning models. At the same time, responsible deployment requires careful monitoring, human oversight, and evaluation under changing market conditions. We release the framework as a research contribution toward more interpretable multimodal financial reasoning rather than as an automated trading system.


\end{document}